\documentstyle{article}
 \renewenvironment{abstract}{\section*{Abstract}\small}{}
       \newtheorem{definition}{Definition}[section]
       \newtheorem{example}[definition]{Example}
       
       \newtheorem{theorem}[definition]{Theorem}
       
       \newtheorem{lemma}[definition]{Lemma}

       \newtheorem{proposition}[definition]{Proposition}
       
       \newtheorem{remark}[definition]{Remark}

       \newtheorem{corollary}[definition]{Corollary}
       \makeatletter
       \renewcommand{\@begintheorem}[2]{ 
  \trivlist\item[\hskip\labelsep{\bf #1\ #2}]}
       \renewcommand{\@opargbegintheorem}[3]{\trivlist
         \item[\hskip \labelsep{\bf #1\ #2\ (#3)}]}
       \makeatother
       \newtheorem{proof}{Proof}

\def \onln {\mbox{${\cal O}{\cal N}{\cal L}_n$}}
\newcommand{\citeyear}{\cite} 

\newcommand{\thm}{\begin{theorem}}
\newcommand{\lem}{\begin{lemma}}
\newcommand{\pro}{\begin{proposition}}
\newcommand{\dfn}{\begin{definition}}
\newcommand{\rem}{\begin{remark}}
\newcommand{\xam}{\begin{example}}
\newcommand{\cor}{\begin{corollary}}
\newcommand{\prf}{\begin{proof} }
\newcommand{\ethm}{\end{theorem}}
\newcommand{\elem}{\end{lemma}}
\newcommand{\epro}{\end{proposition}}
\newcommand{\edfn}{\end{definition}}
\newcommand{\erem}{\end{remark}}
\newcommand{\exam}{\end{example}}
\newcommand{\ecor}{\end{corollary}}
\newcommand{\eprf}{\end{proof}}
\newcommand{\beqn}{\begin{equation}}
\newcommand{\eeqn}{\end{equation}}

\newcommand{\bbox}{\vrule height7pt width4pt depth1pt}

\newcommand{\sat}{\models}

\newcommand{\rimp}{\Rightarrow}

\newcommand{\dimp}{\Leftrightarrow}

\newcommand{\band}{\bigwedge}
\newcommand{\union}{\cup}
\newcommand{\inter}{\cap}

\newfont{\sqi}{cmssqi8}

\renewcommand{\phi}{\varphi}

\newcommand{\K}{{\cal K}}

\newcommand{\N}{{\cal N}}

\newcommand{\ie}{i.e.,~}

\newcommand{\cf}{cf.~}

\newcommand{\respc}{resp.,\ }

\newcommand{\wrt}{with respect to~}

\newcommand{\ol}{\setlength{\itemsep}{0pt}\begin{enumerate}}
\newcommand{\eol}{\end{enumerate}\setlength{\itemsep}{-\parsep}}
\newcommand{\ul}{\setlength{\itemsep}{0pt}\begin{itemize}}
\newcommand{\dl}{\setlength{\itemsep}{0pt}\begin{description}}
\newcommand{\edl}{\end{description}\setlength{\itemsep}{-\parsep}}
\newcommand{\eul}{\end{itemize}\setlength{\itemsep}{-\parsep}}

\newcommand{\true}{\mbox{{\it true}}}
\newcommand{\false}{\mbox{{\it false}}}
 
\newcommand{\Dax}{{\rm KD45}_n}

\newcommand{\commentout}[1]{}

\newcommand{\bi}{\begin{itemize}}
\newcommand{\ei}{\end{itemize}}
\newcommand{\be}{\begin{enumerate}}
\newcommand{\ee}{\end{enumerate}}

\newcommand{\deleted}[1]{{}}

\def \md {{\mathrel|\joinrel=}}
\def \nmd {{\mathrel|\joinrel\neq}}
\def \And {\bigwedge}

\def \onl {\mbox{${\cal O}{\cal N}{\cal L}$}}
\def \onln {\mbox{${\cal O}{\cal N}{\cal L}_n$}}
\def \onlnminus {\mbox{${\cal O}{\cal N}{\cal L}_n^-$}}
\def \KFFn {\mbox{{\rm K45}$_n$}}

\def \Gammabar {\overline{\Gamma}}

\def \Mc {\mbox{$M^c$}}

\def\GammaLth{{\Gamma_L^\theta}}
\def\GammaNth{{\Gamma_N^\theta}}
\def\GammaLthp{{\Gamma_L^{\theta'}}}
\def\GammaNthp{{\Gamma_N^{\theta'}}}
\def \onlp {\mbox{${\cal O}{\cal N}{\cal L}_n^+$}}
\newcommand{\OLNm}{{\cal ONL}_n^-}

\newcommand{\satx}{\sat^x}
\newcommand{\conse}{\sat^e}
\newcommand{\econsequence}{e-consequence}
\newcommand{\Wax}{\mbox{K45$_n$}}
\newcommand{\sWax}{{{\rm K45}_n}}
\newcommand{\obji}{\mbox{{\it obj}}_i}
\newcommand{\objj}{\mbox{{\it obj}}_j}
\newcommand{\Obji}{\mbox{{\it Obj}}_i}
\newcommand{\Objj}{\mbox{{\it Obj}}_j}
\newcommand{\objsi}{\mbox{{\it ob\j}}^+_i}

\newcommand{\Objsi}{\mbox{{\it Ob\j}}^+_i}
\newcommand{\Objsj}{\mbox{{\it Ob\j}}^+_j}
\newcommand{\objei}{\mbox{{\it ob\j}}^e_i}
\newcommand{\Objei}{\mbox{{\it Ob\j}}^e_i}
\newcommand{\subji}{\mbox{{\it subj}}_i}

\newcommand{\mdcm}{\sat^c}
\newcommand{\Conn}{\mbox{\em Sat}}
\newcommand{\Vall}{\mbox{\em Val}}
\newcommand{\Con}[1]{\mbox{\em Sat($#1$)}}
\newcommand{\Val}[1]{\mbox{\em Val($#1$)}}

\newcommand{\newl}[1]{\mbox{\underline{\it\smash{#1}\vphantom{\lower.1ex\hbox{x}}}}}

\newcommand{\onlnzero}{\mbox{${\cal O}{\cal N}{\cal L}_n^0$}}
\newcommand{\onlnk}{\mbox{${\cal O}{\cal N}{\cal L}_n^k$}}
\newcommand{\onlnkplusone}{\mbox{${\cal O}{\cal N}{\cal L}_n^{k+1}$}}
\newcommand{\AXnew}{\mbox{AX$^*$}}
\newcommand{\Afivenewone}{\mbox{\bf A5$_n^1$}}
\newcommand{\Afivenewtwo}{\mbox{\bf A5$_n^2$}}
\newcommand{\Afivenewk}{\mbox{\bf A5$_n^k$}}
\newcommand{\Afivenewkplusone}{\mbox{\bf A5$_n^{k+1}$}}

\begin{document}
\begin{titlepage}
\title{\bf Multi-Agent Only Knowing}
\author{{\bf Joseph Y. Halpern}\\
Department of Computer Science\\
Cornell University\\
Ithaca, NY 14850\\
halpern@cs.cornell.edu\\
http://www.cs.cornell.edu/home/halpern\\
\and {\bf Gerhard Lakemeyer}\\
Department of Computer Science\\
Aachen University of Technology\\
D-52056 Aachen, Germany\\
gerhard@cs.rwth-aachen.de\\
http://www-i5.informatik.rwth-aachen.de/gerhard}
\maketitle
\thispagestyle{empty}

\begin{abstract}
Levesque introduced a notion of ``only knowing'',
with the goal of capturing certain types of nonmonotonic reasoning.
Levesque's logic dealt with only the case of a single agent.
Recently, both Halpern and
Lakemeyer independently attempted to extend Levesque's
logic to the multi-agent case.  Although there are a number of
similarities in their approaches, there are some significant
differences.  In this paper, we reexamine the notion of only knowing,
going back to first principles.  In the process,
we simplify Levesque's completeness proof, and
point out some
problems with the earlier definitions.  This leads us to reconsider
what the properties of only knowing ought to be.  We provide an axiom
system that captures our desiderata, and show that it has a semantics
that corresponds to it.  The axiom system has an added feature of
interest: it includes a modal operator for satisfiability, and thus
provides a complete axiomatization for
satisfiability in the logic K45.
\end{abstract}
\end{titlepage}
\section{Introduction}
\label{intro}
Levesque \citeyear{Lev5} introduced a notion of ``only knowing'',
with the goal of capturing certain types of nonmonotonic reasoning.
In particular, he hoped to capture the type of reasoning that says
``If all I know is that Tweety is a bird, and that birds typically fly,
then I can conclude that Tweety flies''.%
\footnote{The reader should feel free to substitute ``believe''
anywhere we say ``know''.  Indeed, the formal logic that we use, which is
based on the modal logic K45, is more typically viewed as a
logic of belief rather than knowledge.}
Levesque's logic dealt only with the case of a single agent.
It is clear that in many applications of such nonmonotonic reasoning,
there are several agents in the picture.
For example, it may be the
case that all Jack knows about Jill is that Jill knows that Tweety is a
bird and that birds typically fly.  Jack may then want to conclude that
Jill knows that Tweety flies.%

Recently,
each of us \cite{Hal11,Lakemeyer93}
independently attempted to extend
Levesque's
logic to the multi-agent case.  Although there are a number of
similarities in the approaches, there are some significant
differences.  In this paper, we reexamine the notion of only knowing,
going back to first principles.  In the process, we point out some
problems with both of the earlier definitions.  This leads us
to consider what the properties of only knowing ought to be.
We provide an axiom system that captures all our desiderata,
and show that it has a semantics that corresponds to it.  The axiom
system has an added feature of interest: it involves enriching the
language with a modal operator for satisfiability, and thus provides
an axiomatization for satisfiability in K45.
Unfortunately,
the semantics corresponding to this axiomatization
is not
as natural as we might like.
It remains an open question whether
there is a natural semantics for only knowing that corresponds to this
axiomatization.

The rest of this paper is organized as follows.  In the next section,
we review the basic ideas of Levesque's logic and provide an
alternative semantics.
The use of the alternative semantics leads to a simplification of
Levesque's completeness proof. In
Section~\ref{canonical-model}, we review Lakemeyer's
approach, which we call the {\em
canonical-model} approach, and discuss some of its strengths and
weaknesses.
In
Section~\ref{tree-approach}, we go through the same process for
Halpern's
approach.
In Section~\ref{synthesis}, we consider our new approach.
Much of the discussion in Sections~\ref{canonical-model},
\ref{tree-approach}, and~\ref{synthesis} is carried out in terms of
three critical properties of Levesque's approach which we cull out in
Section~\ref{review}.  In particular, we examine to what extent each of
the approaches satisfies these properties.
In Section~\ref{apps}, we show how the logic can be used, and discuss
its relationship to Moore's {\em autoepistemic logic} \cite{Moore85}.
Levesque showed that the single-agent version of his logic of only
knowing was closely connected to autoepistemic logic.  We extend his
result to the multi-agent case.
We conclude in
Section~\ref{discussion} with some discussion of only knowing.
\section{Levesque's Logic of Only Knowing}\label{review}
We begin by reconsidering Levesque's definition.
Let $\Phi$ be a set of primitive propositions.
Let $\onl(\Phi)$ be a propositional
modal language
formed by starting with the primitive propositions in $\Phi$, and
closing off under the classical operators $\lnot$ and $\lor$ and
two modalities, $L$ and $N$.
We omit the $\Phi$ whenever it is clear from context or not relevant
to the discussion.
We freely use other connectives like $\land$,
$\rimp$, and $\dimp$ as syntactic abbreviations of the usual kind.
In addition, we take $O \alpha$ to be an abbreviation for $L \alpha
\land N \neg \alpha$.
Here $L\alpha$ should be read as ``the agent knows or believes
(at least)
$\alpha$'', $N \alpha$ should be read as ``the agent believes at most
$\neg \alpha$'' (so that $N \neg \alpha$ is ``the agent believes at most
$\alpha$'')  and
$O\alpha$  should be read as ``the agent
knows only $\alpha$''.
We define an \newl{objective\/} formula to be a propositional formula
(\ie a formula with no modal operators),
a \newl{subjective\/} formula to be
a Boolean combination of formulas of the form $L\phi$ or $N \phi$,
and a \newl{basic} formula to be a formula which does not
mention $N$.

Levesque gave semantics to knowing and only knowing using the standard
possible-worlds approach.  In the single-agent case, we can identify a
\newl{situation\/} with a pair $(W,w)$, where $w$ is a possible world
(represented as a truth assignment to the primitive propositions)
and $W$ consists of a set of possible worlds.
Intuitively, $W$ is the set of worlds which the agent considers (epistemically)
possible, and $w$ describes the real world.
We do not require that $w \in W$ or that $W
\ne \emptyset$.%
\footnote{By requiring that $W$ is nonempty, we get the modal logic
KD45; by requiring that $w \in W$, we get S5.}
As usual, we say that the agent knows (at least) $\alpha$ if $\alpha$
is true in all the worlds that the agent considers possible.
Formally,
the semantics of the modality $L$ and the classical connectives is given
as follows.
%

\begin{tabular}{l}
  $(W,w) \sat p$ if $w \sat p$
if $p$ is a primitive proposition.\\
  $(W,w) \sat \lnot\alpha$ if $(W,w) \nmd \alpha$.\\
  $(W,w) \sat \alpha \lor \beta$ if $(W,w) \sat \alpha$ or $(W,w) \sat
\beta$.\\
  $(W,w) \sat L \alpha$ if $(W,w') \sat \alpha$ for all $w' \in W$.\\
\end{tabular}

Notice that if $L\alpha$ holds, then the agent may know more than
$\alpha$.  For example, $Lp$ does not preclude $L(p \land q)$ from
holding.
This is why
we should think of $L\alpha$ as saying that the agent knows
{\em at least\/} $\alpha$.

It is well-known that this logic is characterized by the axiom system
K45.  For convenience, we describe K45 here:
\begin{description}
\item[Axioms:]\ \\[1ex]
\begin{tabular}{ll}
{\bf P}. & All instances of axioms of propositional logic.\\
{\bf K}. & $(L\phi \land L(\phi \rimp \psi)) \rimp L\psi$.\\
{\bf 4}. & $L \phi \rimp L L \phi$.\\
{\bf 5}. & $\neg L\phi \rimp L \neg L\phi$.\\
\end{tabular}
\item [Inference Rules:]\ \\[1ex]
\begin{tabular}{ll}
{\bf R1}. & From $\phi$ and $\phi \rimp \psi$ infer $\psi$.\\
{\bf R2}. & From $\phi$ infer $L\phi$.
\end{tabular}
\end{description}
The axioms {\bf 4} and {\bf 5} are called the {\em positive
introspection axiom\/} and {\em negative introspection axiom},
respectively.  They are appropriate for agents that are sufficiently
introspective so that they know what they know and do not know.


How do we give precise semantics to $N$?  That is, when should we say
that $(W,w) \sat N \beta$?
Intuitively, $N \beta$ is
true if $\beta $ is true at all the worlds that the agent does {\em not\/}
consider possible.
It seems fairly clear from the intuition
that we need to evaluate the truth of $\beta$ in worlds
$w' \notin W$,\footnote{Note that, since we defined worlds extensionally as truth
assignments, the set of impossible worlds is well-defined and
fixed for a given $W$.}
since these are the worlds that the agent considers
impossible in $(W,w)$.  But if $\beta$ is a complicated formula
involving nested $L$ operators, then we cannot simply evaluate the truth
of $\beta$ at a world $w'$.  We need to have a set of worlds too.
In fact, the set of possible worlds we use is still $W$.  That is, while
evaluating the truth of $\beta$ in the impossible worlds, the agent
keeps the set of worlds he considers possible fixed.
Formally,
we define
$$\mbox{$(W,w) \sat N \alpha$ if $(W,w') \sat \alpha$ for all $w' \notin
W$.}$$

Let us stress three important features of this definition.
\begin{itemize}
\item First, as we have
already observed, the set of possibilities is kept fixed when
we evaluate $N\alpha$.
\item Second, the set of \newl{conceivable worlds\/}---%
the union of the set of ``possible'' worlds considered when evaluating
$L$ and the set of ``impossible'' worlds considered when evaluating $N$%
---is fixed,
independent of the situation $(W,w)$; it is always the set of all truth
assignments.
\item 
Finally, for every set of conceivable worlds, there is a model where
that set is precisely the set of worlds that the agent considers
possible.
\end{itemize}


Roughly speaking, the first property is what is required for A4, while
the second property is what is required for A5.
 The intent of the third property is to make it
possible that any objective formula can be ``all you know.''
As we shall see in Section~\ref{synthesis}, in a precise sense, the
third property is somewhat stronger than we actually need.
All we really need is that for any
objective formulas the set of all
worlds satisfying these formulas can be considered possible.
We return to these properties for guidance when we
discuss possible ways of extending Levesque's semantics to the
multi-agent case.

Since $O \alpha$ is an abbreviation for $L \alpha \land N \neg \alpha$,
we have that
\[
\mbox{$(W,w) \sat O \alpha$ if
                   for all worlds $w'$, $w'\in W$ iff $(W,w')\sat\alpha$.}
\]



As it stands, the semantics has the somewhat odd property that there are
situations that agree on all basic beliefs yet disagree on what is only
believed. As pointed out by Levesque~\citeyear{Lev5}, the problem is that there
are far too many sets of worlds than there are basic belief sets.  In order to
find a perfect match between the sets of basic beliefs an agent may hold and
sets of worlds, Levesque introduces what he calls \newl{maximal\/} sets of
worlds. In essence, a maximal set is the largest set in the sense that adding
any other world to it would change the agent's basic beliefs. Furthermore, every
set of worlds can be extended to a unique maximal set of worlds.  It is well
known that in the logic K45, an agent's beliefs are completely determined by his
beliefs about objective formulas (see, for example, \cite{HM4} for a proof).
Thus, we define a maximal set as follows:

\dfn If $W$ is a set of worlds, let
$$W^+ = \{ w\;|\; \mbox{for all objective formulas $\phi$, if
$(W,w)\md L\phi$ then
$(W,w)\md\phi$} \}.$$
$W$ is called {\em maximal\/} iff $W = W^+$.
\edfn
Levesque defines validity and satisfiability with
respect to {\em maximal\/}
sets only. In particular, a formula $\alpha$ is valid iff for every
maximal set of worlds $W$ and every world $w \in W$, we have
$(W,w)\md\alpha$.

We end this
review of Levesque's logic by presenting (a slight variant
of) his proof theory.
\begin{description}
\item [Axioms:]\ \\[1ex]
\begin{tabular}{ll}
\label{axioms-of-ol}
{\bf A1}. & All instances of axioms of propositional logic.\\
{\bf A2}. & $L(\alpha \rimp \beta) \rimp (L\alpha \rimp L\beta)$.\\
{\bf A3}. & $N(\alpha \rimp \beta) \rimp (N\alpha \rimp N\beta)$.\\
{\bf A4}. & $\sigma \rimp L\sigma \land N\sigma$ for every subjective
formula $\sigma$.\\
{\bf A5}. & $N\alpha\rimp\lnot L\alpha$ if $\neg \alpha$ is a
propositionally consistent
 objective
formula.\\
\end{tabular}
\item [Inference Rules:]\ \\[1ex]
\begin{tabular}{ll}
{\bf MP}. & From $\alpha$ and $\alpha\rimp\beta$ infer $\beta$.\\
{\bf Nec}. & From $\alpha$ infer $L\alpha$ and $N\alpha$.
\end{tabular}
\end{description}
Axioms {\bf A2}--{\bf A4} tell us that
that $L$ and $N$
separately have all the properties of K45-operators.
Actually, {\bf A4} tells us more; it says that
$L$ and $N$ are mutually introspective, so that, for example, $L \phi
\rimp NL \phi$ is valid.
Perhaps the most interesting axiom is {\bf A5},
which gives only-knowing its desired properties.
Its soundness depends on the fact that
the union of the set of worlds considered when evaluating $L$ and
the set of worlds considered when evaluating $N$ is the set of all
conceivable worlds.%
\footnote{Note that, while unusual, the
axiom schema {\bf A5} is
recursive,
since consistency of formulas in classical
propositional logic is decidable.
Hence the axioms themselves are {\em recursive}.
As noted in~\cite{Lev5}, this is
a problem in the first-order case, however.
In fact,
Levesque's proof theory for the first-order
version of his logic was recently shown to be incomplete~\cite{HL95}.
}
\thm\label{levesque} {\rm \cite{Lev5}}
If $\Phi$ is infinite, then
Levesque's axiomatization is sound and complete
for the language $\onl(\Phi)$
with respect to Levesque's semantics.
\ethm

As we shall see, the assumption that there are infinitely many primitive
propositions in $\Phi$ is crucial for Levesque's completeness result.
Extra axioms are required if $\Phi$ is finite.  In addition,
it is interesting to note that the assumption that $L$ and $N$ are
interpreted with respect to complementary sets of worlds is not forced
by the axioms. In particular,
for the soundness of Axiom {\bf A5}, it suffices that
the sets considered for $L$ and $N$ cover all conceivable worlds;
they may overlap.
The following semantics makes this
precise.

Define an \newl{extended situation} to be a triple $(W_L,W_N,w)$,
where $W_L$ and $W_N$ are sets of worlds (truth assignments) such that
$W_L \union W_N$ consists of all truth assignments.
Define a
new satisfaction relation $\satx$ that is exactly like Levesque's
except for $L$- and $N$-formulas.
For them, we have

\begin{tabular}{l}
$(W_L,W_N,w)
\satx L \alpha$ if $(W_L,W_N,w') \satx \alpha$ for all $w' \in W_L$\\
$(W_L,W_N,w)
\satx N \alpha$ if $(W_L,W_N,w') \satx \alpha$ for all $w' \in W_N$.\\
\end{tabular}

Note that $L$ and $N$ are now treated in a completely symmetric way.

\thm\label{Levthm1}
For all $\Phi$,
Levesque's axiomatization is sound and complete
for the language $\onl(\Phi)$
with respect to
$\satx$.
\ethm

\prf
We omit the soundness proof, which is straightforward. Note that for axiom
{\bf A5} to be sound it suffices that $W_L$ and $W_N$ together cover all worlds.  In
particular, it does not matter whether or not the two sets overlap.

To prove completeness, we use the notion of a {\em maximal consistent\/}
set.  Given an arbitrary axiom system $AX$, we say that a formula $\phi$
is {\em consistent with respect to AX\/} if it is not the case that
$AX \vdash \neg \phi$, where,
as usual, we use $\vdash$ to denote provability.
A finite set of formulas $\phi_1, \ldots,
\phi_n$ is consistent with respect to $AX$ if the conjunction $\phi_1
\land \ldots \land \phi_n$ is consistent with respect to $AX$.  An
infinite set of formulas is consistent with respect to $AX$ if every
finite subset of its formulas is consistent with respect to $AX$.
Finally, given a set $F$ of formulas, a {\em maximal consistent\/}
subset of $F$ is a subset $F'$ of $F$ which is consistent with respect
to $AX$ such that any superset of $F'$ is not consistent with respect to
$AX$.

\begin{sloppypar}
In the following, provability, consistency,
and maximal consistency all refer to Levesque's axiom system unless stated
otherwise.
To prove completeness we show that every consistent formula is
satisfiable with respect to $\satx$, using a standard canonical model
construction \cite{HM2,HC}.
\end{sloppypar}

Let $\Gamma_0$ be the set of all maximal consistent sets of formulas in
$\onl(\Phi)$.   For $\theta\in\Gamma_0$, define  $\theta/ L =
  \{\alpha\;|\; L\alpha\in\theta\}$ and
$\theta/ N =   \{\alpha\;|\; N\alpha\in\theta\}$.  We then define
\begin{itemize}
\item $\GammaLth = \{\theta'\in\Gamma_0\;|\; \theta/ L\subseteq \theta'\}$,
\item $\GammaNth = \{\theta'\in\Gamma_0\;|\; \theta/ N\subseteq \theta'\}$.
\end{itemize}

If we view maximal consistent sets as worlds, then $\GammaLth$ and
$\GammaNth$ represent the worlds accessible from $\theta$ for $L$ and $N$,
respectively. The following lemma reflects the fact that $L$ and $N$ are
both fully and mutually introspective (axiom {\bf A4}).

\begin{lemma}
\label{lemma-single-ag-completeness-proof}
If $\theta'\in \GammaLth\union\GammaNth$, then $\GammaLthp = \GammaLth$
and $\GammaNthp = \GammaNth$.
\end{lemma}
\prf
We prove the lemma for $\theta'\in \GammaLth$. The case $\theta'\in \GammaNth$
is completely symmetric.
%
To show that $\GammaLth = \GammaLthp$, it clearly suffices to show
that $\theta/ L = \theta'/ L$.
Let $\alpha\in \theta/ L$. Then $L\alpha\in\theta$ and
also $LL\alpha\in\theta$ by axiom {\bf A4}. Thus $L\alpha\in\theta'$
(since $\theta' \in \GammaLth$ implies that $\theta/L \subseteq \theta'$)
and, hence, $\alpha\in\theta'/ L$.

For the converse, let $\alpha\in \theta'/ L$.  Thus,
$L\alpha\in\theta'$.
Assume that $\alpha\not\in \theta/ L$. Then $\lnot L\alpha\in\theta$
(since $\theta$ is a maximal consistent set)
and, therefore, $L\lnot L\alpha\in\theta$, from which
$\lnot L\alpha\in\theta'$ follows, a contradiction.

The proof that $\GammaNthp = \GammaNth$ proceeds the same way, that is,
we show that $\theta/ N = \theta'/ N$.
Let $\alpha\in \theta/ N$. Then $N\alpha\in\theta$ and
also $LN\alpha\in\theta$ by axiom {\bf A4}. Hence $N\alpha\in\theta'$,
so $\alpha\in \theta'/ N$.

For the converse, let $\alpha\in \theta'/ N$.  Thus,
$N\alpha\in\theta'$. Assume that $\alpha\not\in \theta/ N$.
Then $\lnot N\alpha\in\theta$
and also $L\lnot N\alpha\in\theta$,
from which $\lnot N\alpha\in\theta'$ follows, a contradiction.
\eprf

\def\WLth{{W_L^\theta}}
\def\WNth{{W_N^\theta}}
\def\WLthp{{W_L^{\theta'}}}
\def\WNthp{{W_N^{\theta'}}}
In traditional completeness proofs using maximal consistent sets
(see, for example, \cite{HM2,HC}), a situation is constructed whose
worlds consists of all maximal consistent sets.  Here, we must be a
little more careful.

We say that a maximal consistent set $\theta$ contains a truth assignment
$w$ if for all atomic formulas $p$, we have $w\sat p$ iff $p\in\theta$.
Clearly a maximal consistent set $\theta$ contains exactly one world;
we denote this world by $w_\theta$.
For $\theta\in\Gamma_0$,
let
$\WLth = \{ w_{\theta'} \;|\; \theta'\in\GammaLth\}$ and
$\WNth = \{ w_{\theta'} \;|\; \theta'\in\GammaNth\}$.
\begin{lemma}
\label{completeness-max-sets-extended-sits}
\begin{enumerate}
\item[(a)] $(\WLth,\WNth,w_\theta)$ is an extended situation.
\item[(b)] For all $\alpha$, we have $\alpha\in\theta$ iff
$(\WLth,\WNth,w_\theta)\satx \alpha$. \end{enumerate}
\end{lemma}
\prf
For part (a),
to show that $(\WLth,\WNth,w_\theta)$ is an extended situation, we
must show that $\WLth\union\WNth$ consists of all truth assignments.  By
way of contradiction, suppose there is a truth assignment $w$ not in
$\WLth\union\WNth$.
Let $F_w = \{p \in \Phi \;|\; w \sat p\} \union
\{\neg p \;|\; p \in \Phi, \; w \sat \neg p\}$.
$F_w \union \theta/L$ cannot be consistent, for otherwise there would be
some $\theta'
\in \GammaLth$ that contains $F_w$, which would mean that $w \in
\WLth$.  Similarly $F_w \union \theta/N$ cannot be consistent.
Thus, there must be formulas $\phi_1, \phi_2, \phi_3, \phi_4$ such that
$\phi_1$ and $\phi_2$ are both conjunctions of a finite number of
formulas
in $F_w$, $\phi_3$ is the conjunction of a finite number of formulas in
$\theta/L$, and $\phi_4$ is the conjunction of a finite number of
formulas in $\theta/N$, and both $\phi_1 \land \phi_3$ and $\phi_2 \land
\phi_4$ are inconsistent.  Thus, we have $\vdash \phi_3 \rimp \neg
\phi_1$ and $\vdash \phi_4 \rimp \neg \phi_2$.  Using standard modal
reasoning ({\bf A2}, {\bf A3}, and {\bf Nec}), we have
$\vdash L \phi_3 \rimp L \neg \phi_1$ and $N \phi_4 \rimp N \neg
\phi_2$.  Since $L \psi \in \theta$ for each conjunct $\psi$ of
$\phi_3$, standard modal reasoning shows that $L \phi_3 \in \theta$.
Similarly, we have $N \phi_4 \in \theta$.  Since $\theta$ is a maximal
consistent set, both $L \neg \phi_1$ and $N \neg \phi_2$ are in
$\theta$.  Since $\vdash L \neg \phi_1 \rimp L(\neg \phi_1 \lor \neg
\phi_2)$ and $\vdash N \neg \phi_2 \rimp N(\neg
\phi_1 \lor \neg \phi_2)$, it
follows that both $L (\neg \phi_1 \lor \neg \phi_2)$ and $N (\neg \phi_1
\lor \neg \phi_2)$ are in $\theta$.  But this contradicts {\bf A5},
since $\phi_1 \land \phi_2$ is a propositionally consistent objective
formula.

For part (b),
the proof proceeds by induction on the structure of $\alpha$.
The statement
holds trivially for atomic propositions, conjunctions, and negations.
In the case of $L \alpha$, we proceed by the following chain of
equivalences:

\begin{tabular}{ll}
&$L\alpha\in\theta$\\
iff &for all $\theta'\in\GammaLth$, we have $\alpha\in\theta'$\\
iff &for all $\theta'\in\GammaLth$, we have
$(\WLthp,\WNthp,w_{\theta'})\satx \alpha$ (using the induction
hypothesis)\\
iff &for all $w_{\theta'}\in \WLth$, we have
$(\WLth,\WNth,w_{\theta'})\satx \alpha$
(by Lemma~\ref{lemma-single-ag-completeness-proof})\\
iff
&$(\WLth,\WNth,w_{\theta})\satx L\alpha$.
\end{tabular}

\noindent The case $N\alpha$ is completely
symmetric. \eprf

The completeness result now follows easily.
Let $\alpha$ be a consistent formula and $\theta$ a maximal consistent
set of formulas containing $\alpha$.
By Lemma~\ref{completeness-max-sets-extended-sits},
$(\WLth,\WNth,w_{\theta})\satx \alpha$.
\eprf

Levesque considered only maximal sets in his definition of validity.
In fact, this restriction has no effect on the notion of validity.

\cor
A formula $\alpha \in \onl(\Phi)$ is valid iff $(W,w)\sat\alpha$ for all
situations $(W,w)$ (including nonmaximal $W$).
\ecor
\prf
If $\Phi$ is finite, it is easy to check that $W^+ = W$ for all sets
$W$, so the result is trivially true if $\Phi$ is finite.  So suppose
$\Phi$ is infinite.
Notice that each situation $(W,w)$ corresponds to an extended situation
$(W_L,W_N,w)$, where $W_L = W$ and $W_N$ is the complement of $W$.
Let us call such an $(W_L,W_N,w)$ an {\em extended complementary
situation}.
Theorems~\ref{levesque} and~\ref{Levthm1} together imply
the valid formulas obtained when considering all extended
situations remain the same when we restrict ourselves to
complementary situations with maximal $W_L$.
The corollary then follows from the fact that the set of all extended
situations
properly includes the set of all extended complementary situations, which
in turn includes the set of all extended complementary situations with
maximal $W_L$.
\eprf

\deleted{
We can actually prove completeness of Levesque's axioms for $\satx$
directly, using a standard canonical model construction (\cf
Section~\ref{canonical-model}).  Moreover, as we show in the full
paper, we can use our approach
to give a simpler completeness proof for Levesque's semantics than
the one given by Levesque.  The key idea is to find a model satisfying
a formula where $W_L$ and $W_N$ may overlap, and then
use the truth assignment to a primitive proposition not mentioned
in the formula to force $W_L$ and $W_N$ to be disjoint.  We know
that there is bound to be a primitive proposition not mentioned in
the formula, given the assumption that the set of primitive
propositions is infinite.
} 

As Theorem~\ref{Levthm1} shows, for the $\satx$ semantics, Levesque's
axioms are sound and complete for all sets $\Phi$ of primitive
propositions.  On the other hand,
as we said earlier, Levesque's completeness proof (with respect
to his semantics) depends crucially on the fact that $\Phi$ is infinite.
If $\Phi$ is finite, Levesque's axioms are still sound with respect to
his semantics, but they are no longer complete.
For example,
if $\Phi=\{p\}$,
then $\neg L \neg p \rimp N \neg p$ would be valid under $\sat$; this
does not follow from the axioms given above.
In fact, for each finite set $\Phi$ of primitive propositions, we can
find a new axiom scheme that, taken together with the previous axioms,
gives a complete axiomatization for $\onl(\Phi)$ for Levesque's
semantics if $\Phi$ is finite.%
\footnote{This was also the  situation
for the logic
considered in \cite{FHV2}.  In that paper, a simple axiomatization was
provided for the case where $\Phi$  was
infinite; for each finite $\Phi$, an extra axiom was needed (that
depended on $\Phi$).
}
The new axiom, which subsumes axiom {\bf A5}, allows us to reduce
formulas involving $N$
formulas involving only $L$.

Note that worlds, which are truth assignments to the
primitive propositions $\Phi$, are themselves finite if $\Phi$ is finite.
Hence
we can identify a world $w$ with the conjunction of all literals over $\Phi$
that are true at $w$. For example, if $\Phi = \{p,q\}$ and
$w$ makes $p$ true and $q$ false, then we identify $w$ with
$p\land\lnot q$.
For any objective formula $\alpha$, let $W_{\alpha,\Phi}$ be
the set of all worlds (over the primitive propositions $\Phi$) that
satisfy $\alpha$.
The axiom system
AX$_{\Phi}$
is then obtained from Levesque's system
by replacing {\bf A5} by the following axiom:

\begin{tabular}{ll}
{\bf A5$_{\Phi}$}.  $N\alpha \equiv \And_{w\in W_{\lnot\alpha,\Phi}}
\lnot L\lnot w$
if $\neg \alpha$ is a propositionally consistent  objective formula.
\end{tabular}

The axiom is easily seen to be sound since it merely expresses
that $N\alpha$ holds at $W$ just in case $W$
contains all worlds that satisfy $\lnot\alpha$.
Note that this property depends only on the fact that $L$ and $N$ are
defined with respect to complementary sets of worlds and, hence, also holds
in the case of infinite $\Phi$. However, it is only in
the finite case that we can express this axiomatically.
Completeness is also very easy to establish.
Levesque~\citeyear{Lev5}
showed that in his system, even without {\bf A5}, every formula is
provably
equivalent to one without nested modalities. With {\bf A5$_{\Phi}$}, we
then obtain an equivalent formula that does not mention $N$.
In other words, given a formula consistent with respect to AX$_{\Phi}$,
a satisfying model can be constructed with the usual technique
for $K45$ alone.

\thm
AX$_{\Phi}$ is sound and complete
for the language $\onl(\Phi)$
with respect to
Levesque's semantics, if $\Phi$ is finite.
\ethm

\section{The Canonical-Model Approach}\label{canonical-model}

How do we extend our intuitions about only knowing to the multi-agent case?
First we extend
the language $\onl(\Phi)$ to the case of many agents.  That is, we now
consider
a language $\onln(\Phi)$, which is just like $\onl$ except that there
are modalities $L_i$ and $N_i$ for each agent $i$, $1\le i\le n$, for
some fixed $n$.
In the remainder of the paper, we omit the $\Phi$, just writing $\onl$
and $\onln$, since the set of primitive propositions does not play a
significant role.
By analogy with the single-agent
case,
we call a formula {\em basic} if it does not mention any of the operators $N_i$
($i = 1, \ldots, n$) and
\newl{$i$-subjective\/} if it is  a Boolean combination of formulas
of the form $L_i \phi$ and $N_i \phi$.   What should be the analogue of
an objective formula? It clearly is more than just a propositional
formula.  From agent 1's point of view, a formula like $L_2 p$ or even
$L_2 L_1 p$ is just as ``objective'' as a propositional formula.
We define a formula to be \newl{$i$-objective\/} if it is a Boolean
combination of primitive propositions and
formulas of the form $L_j\phi$ and $N_j \phi$, $j \ne i$,
where $\phi$ is arbitrary.
Thus,
$q \land N_2 L_1p$ is 1-objective, but
$L_1 p$ and $q \land L_1p$ are not.  The $i$-objective formulas true at
a world can be thought of as characterizing what is true apart from
agent $i$'s subjective knowledge of the world.

The standard
model here is to have a {\em Kripke structure\/} with worlds and
accessibility
relations that describe what worlds the agents consider possible in
each world.  Formally, a (Kripke) structure
or model
is a tuple $M = (W,\pi,
\K_1, \ldots, \K_n)$, where $W$ is a set of
worlds,\footnote{Note that here $W$ denotes the set of {\em all}
worlds of the particular model $M$,
not just the (epistemically) possible ones as in Levesque's
logic.}
$\pi$ associates with each world a truth
assignment to the primitive propositions, and
$\K_i$ is agent $i$'s accessibility
relation.
Given such a Kripke structure $M$,
let $\K_i^M(w)
= \{w': (w,w') \in \K_i\}$.%
\footnote{We use the superscript $M$ since we shall later need to talk
about the $\K_i$ relations in more than one model at the same time.}
$\K_i^M(w)$ is the set of worlds that
agent $i$ considers possible at $w$ in structure $M$.
As usual, we define
$$\mbox{$(M,w) \sat L_i \alpha$ if $(M,w') \sat \alpha$ for all $w' \in
\K_i^M(w)$.}$$
We focus on structures where the accessibility relations are Euclidean
and transitive,
where a relation $R$ on $W$ is Euclidean
if $(u,v) \in R$ and $(u,w) \in R$ implies that $(v,w) \in R$, and
$R$ is transitive if $(u,v) \in R$ and $(v,w) \in R$ implies that $(u,w)
\in R$.  We call such structures $\Wax$-structures.
It is well known \cite{Chellas,HM2} that these assumptions
are precisely what is required to get belief to obey the K45 axioms
(generalized to $n$ agents).
We say that a formula consistent with these
axioms is $\Wax$-consistent.
An infinite set of formulas is said to be $\Wax$-consistent if
the conjunction of the formulas in
every one of its finite subsets is $\Wax$-consistent.

Now the question is how to define the modal operator $N_i$.
The problem in the multi-agent case is that we can
no longer identify a possible world with a truth assignment.
In the single-agent case, knowing the set of truth assignments that the
agent considers possible completely determines his knowledge.  This is
no longer true in the multi-agent case.  Somehow we must take the
accessibility relations into account.
A general semantics for an $N$-like operator was first given by
Humberstone~\citeyear{Humberstone} and later by Ben-David and
Gafni~\citeyear{BenDavid}.
In this approach, the semantics of $N_i$ is given as follows:
$$\mbox{$(M,w) \sat N_i \alpha$ if $(M,w') \sat \alpha$ for all $w'
\in W- \K_i^M(w)$}.$$
The problem with this definition is that it misses out on the intuition
that when evaluating $N_i \alpha$, we keep the set of worlds that agent
$i$ considers possible fixed.  If $w' \in W - \K_i^M(w)$, there is
certainly no reason to believe that $\K_i^M(w) = \K_i^M(w')$.

One approach to solving this problem is as follows:
If $w$ and $w'$ are two worlds in $M$, we write \newl{$w \approx_i w'$} if
$\K_i^M(w) = \K_i^M(w')$, \ie if $w$ and $w'$ agree on the possible
worlds according to agent $i$.
We then define
$$\mbox{$(M,w) \sat N_i \alpha$ if $(M,w') \sat \alpha$ for all $w'$
such that $w' \in W - \K_i^M(w)$ and $w \approx_i w'$.}$$

While this definition does capture the first of Levesque's properties,
it does not capture the second.  To see the problem, suppose we have
only one agent and a structure $M$ with only one possible world $w$.
Suppose that $(w,w) \in \K_1^M$ and $p$ is true at $w$.  Then it
is easy to see that $(M,w) \sat L_1 p \land N_1 p$, contradicting axiom
{\bf A5}.
The problem is that since the structure has only one world
and it is in $\K_1^M(w)$,  there are no worlds in $W - \K_1^M(w)$.
Thus, $N_1 p$ is vacuously true.
Intuitively, there just aren't enough
``impossible'' worlds in this case;
the set of conceivable
worlds is not independent of the model.
To deal with this problem,
we focus attention on one particular model, the
{\em canonical model},
which intuitively has ``enough'' worlds.
Its worlds consist of all the maximal consistent subsets of basic
formulas.  (Recall that maximal consistent sets were defined in the
proof of Theorem~\ref{Levthm1}.)  Thus, in some sense, the canonical
model has as many worlds as possible.

\dfn The {\em canonical model\/} (for $\Wax$) $M^c = ( W^c, \pi^c,
\K_1^c,\ldots, \K_n^c )$ is defined as follows:
\begin{itemize}
  \item $W^c = \{ w\;|\; \mbox{$w$ is a maximal consistent set of basic
formulas \wrt K45$_n$}\}$,
  \item for all primitive propositions
  $p$ and $w\in W^c$, $\pi(w)(p) =$ {\bf true}
  iff $p\in w$,
  \item $(w,w') \in \K_i^c$ iff $w/L_i \subseteq w'$, where
  $w/L_i = \{\alpha \;|\; L_i \alpha \in w\}$.
\end{itemize}
\end{definition}
\deleted{
Using standard techniques (see, for example, \cite{Chellas,HM2}), we
can show that
\thm\label{canonicalmodel}
$M^c$ is a $\Wax$ structure, and for each maximal
consistent set $w$, we have
$(M^c,w) \sat \phi \mbox{ iff } \phi \in w$. \ethm
This shows that every satisfiable formula is satisfiable in some state
in $M^c$.

%
} 

Validity in the canonical-model approach is defined with respect to
the canonical model only. More precisely, a formula $\alpha$ is said to be
{\em valid in the canonical-model approach}, denoted
$\mdcm\alpha$, iff
$M^c \md \alpha$, that is, if
for all worlds $w$ in the canonical model
we have $(M^c,w)\md \alpha$.

\deleted{
Note that, in contrast to Levesque's approach, there is no need to consider
maximal sets of worlds, since the canonical model by construction provides a
perfect match between the sets of basic beliefs that an agent may hold and the
sets of possible worlds.
}

We clearly cannot use the canonical model in a practical way.  It  can
be shown that it has uncountably many worlds.  Each of its worlds is
characterized by an infinite set of formulas, so cannot be described
easily.   Moreover, in general, both $\K_i^c(w)$
and $W^c - \K_i^c(w)$ are infinite, so we cannot compute whether $L_i
\phi$ or $N_i \phi$ holds at a given world.  Thus, our interest in the
canonical model is mainly to understand whether it gives reasonable
semantics to the $N_i$ operator.

We start by arguing that,
for an appropriate notion of ``possibility'' and ``conceivability'',
this semantics satisfies the first two of
Levesque's properties.
\deleted{
That is, we want to show that under an
appropriate notion of ``conceivable state'',
the set of conceivable states
for agent $i$ is indeed the same at all worlds.  Notice that we say
``conceivable state'' here rather than ``conceivable world''.}
What then is a conceivable world?
Intuitively, it is an objective state of
affairs from agent $i$'s point of view, which does not include $i$'s
beliefs. In the single-agent case, this is simply a truth assignment.
In the multi-agent case, things are more complicated, since beliefs
of other agents are also part of $i$'s objective world.
One way of characterizing a state of affairs from $i$'s point of view is
by the set of $i$-objective formulas that are true at a particular world.
For
technical reasons, in this section we restrict even further to the
$i$-objective {\em basic\/} formulas---%
that is, those formulas that do not mention any of the modal operators
$N_j$, $j = 1, \ldots, n$---%
that are true.
If we assume that the basic formulas determine all the
other formulas,
which can be shown to be true in the single-agent case, and under this
semantics for the multi-agent case,
then it is arguably reasonable to restrict to basic formulas.
However,
as we
shall see in Section~\ref{tree-approach}, it is not clear that this
restriction is appropriate,
although we make it for now.

\dfn
Given a situation $(M,w)$, let \newl{$\obji(M,w)$} consist of all the
$i$-objective basic formulas that are true at $(M,w)$.
let \newl{$\Obji(M,w)$} = $\{\obji(M,w') \;|\; w'
\in \K_i^M(w)\}$, and let
\newl{$\subji(M,w)$} =
$\{\mbox{basic}\ L_i\alpha\;|\; (M,w)\md L_i\alpha\} \union
              \{\mbox{basic}\ \lnot L_i\alpha\;|\; (M,w)\nmd
L_i\alpha\}$.
\edfn
We take $\obji(M,w)$
to be $i$'s state at $(M,w)$.  Notice that $\obji(M,w)$ is a maximal
consistent set of $i$-objective basic formulas.  For ease of exposition,
we say \newl{$i$-set} from now on rather than ``maximal
set of $i$-objective basic formulas''.  Thus,
the set of conceivable states for agent $i$ is the set of all
$i$-sets.
Notice that the set of conceivable states is
independent of the model.  It is easy to show that this is a
generalization of the
single-agent case, since in the single-agent case the $i$-objective
basic formulas
are just the propositional formulas, and
an $i$-set
can be identified with a truth assignment.

$\Obji(M,w)$ is the set of $i$-sets that agent
$i$ considers possible in situation $(M,w)$;
$\subji(M,w)$ characterizes $i$'s basic beliefs in $(M,w)$.  Notice that
if $\alpha$ is an $i$-objective basic formula, then
$L_i \alpha \in \subji(M,w)$ iff $\alpha$ is in every $i$-set in
$\Obji(M,w)$.

With these definitions, we can
show that the first two of Levesque's properties hold in the canonical
model.
The first property
says that at all worlds $w'$ considered in evaluating a formula of the
form $N_i \phi$ at a world $w$, the set of possible states---that
is, the set $\{\obji(M^c,w'') \;|\; w''
\notin
\K_i^c(w')\}$---is the same
for all $w' \in \K_i^c(w)$.
This is easy to see, since the
only worlds $w'$ we consider are those such that $\K_i^c(w') = \K_i^c(w)$.
The second property says that the union of the set of states associated
with
the worlds used in computing $L_i \phi$ at $w$ and the set of states
associated with the worlds
used in computing $N_i \phi$ at $w$ should consist of all conceivable
states. To show this, we must show that for every world $w$ in the
canonical model, the set $\{\obji(M^c,w') \;|\; w' \approx_i w\}$
consists of all $i$-sets.

To prove this, we need two preliminary lemmas.


\lem
\label{$i$-equivalence-same-as-same-basic-beliefs}
Let $w$ and $w'$ be worlds in $M^c$. Then $w\approx_i w'$ iff
agent $i$ has the same basic beliefs at $w$ and $w'$, that is,
$\subji(M^c,w) = \subji(M^c,w')$.
\elem

\prf
The ``only if'' direction is immediate because $w$ and $w'$ are assumed
to have the same $\K_i$-accessible worlds.
To prove the ``if'' direction,
suppose that $\subji(M^c,w) = \subji(M^c,w')$ but
 $w\not\approx_i w'$. Without loss of generality, there is a world $w^*
\in  \K_i^c(w) - \K_i^c(w')$.
By the definition of the canonical model,
there must be a basic formula $L_i\alpha\in w'$ such that
$\alpha\not\in w^*$. By assumption,
$L_i\alpha\in w$, contradicting the assumption that $w^* \in
\K_i^c(w)$. \eprf

\lem\label{consistency}  Suppose $\Gamma$ consists only of $i$-objective basic
formulas, $\Sigma$ consists only of $i$-subjective basic formulas, and
$\Gamma$ and $\Sigma$ are both $\Wax$-consistent.  Then $\Gamma \union
\Sigma$ is $\Wax$-consistent. \elem

\prf
This follows immediately from part (c) of Proposition~\ref{construction}
below.  \eprf


We can now prove that the set of conceivable states for agent $i$ is the
same at all worlds of the canonical model.
This follows from
the following result.
\thm
\label{all-possibilities-covered}
Let $w\in W^c$. Then for every
$i$-set $\Gamma$
there is exactly one world
$w^*$ such that $\obji(M^c,w^*) = \Gamma$ and $w\approx_i w^*$.
\ethm

\prf
Let $\Sigma = \subji(M^c,w)$.
Since $\Gamma$ consists of $i$-objective basic formulas only,
$\Sigma$ consists of $i$-subjective formulas, and $\Gamma$ and
$\Sigma$ are both $\Wax$-consistent, by
Lemma~\ref{consistency}, $\Gamma \union \Sigma$ is $\Wax$-consistent.
Let $w^*$
be a maximal consistent set that contains $\Gamma\union\Sigma$. Since
$w$ and $w^*$ agree on $\Sigma$, $w\approx_i w^*$
by Lemma~\ref{$i$-equivalence-same-as-same-basic-beliefs}. The uniqueness of $w^*$
follows by a simple induction argument.
\eprf


What about the third property?  This says that every subset of
$i$-sets arises as the set of $i$-sets associated
with the worlds that $i$ considers possible in some situation; that is,
for every set $S$ of $i$-sets, there should be some
situation $(M^c,w)$ such that $S = \Obji(M^c,w)$.  As
we now show, this property does {\em not\/} hold in the canonical model.
We do this by showing that the set of $i$-sets associated with
the worlds considered possible in any situation in the canonical model
all have a particular property we call {\em limit closure}.%
\footnote{This turns out to be closely related to the limit closure
property discussed in \cite{FGHV92}; a detailed comparison would
take us too far afield here though.}
\dfn
\label{limit-closure}
We say that an $i$-set $\Gamma$ is a {\em limit\/} of a set $S$ of
$i$-sets if, for every finite subset $\Delta$ of $\Gamma$, there is a
set $\Gamma' \in S$ such that $\Delta \subset \Gamma'$.  A set $S$ of
$i$-sets is \newl{limit closed\/} if every limit of $S$ is in $S$.
\edfn
%
\lem\label{lc}
For every world $w$ in $M^c$, the set $\Obji(M^c,w)$
is limit closed.
\elem

\prf
Let $w$ be a world in the canonical model and let
$\Gamma$ be an $i$-set which is a limit
of $\Obji(M^c,w)$.  We want to show that $\Gamma \in \Obji(M^c,w)$.  Let
$\Sigma =
\{\mbox{basic } \beta \;|\; L_i \beta \in w\}$.  We claim that $\Gamma
\union \Sigma$ is $\Wax$-consistent.  For suppose not.  Then there must
be a finite subset $\Delta \subseteq \Gamma$ such that $\Delta \union
\Sigma$ is inconsistent.  Since $\Gamma$ is a limit of $\Obji(M^c,w)$,
there
must be some world $w'$ in $\Obji(M^c,w)$ such that $(M^c,w') \sat
\Delta$. By
construction of the canonical model, since $w' \in \K_i^c(w)$, we must
have that $(M^c,w') \sat \Sigma$.  Thus $\Delta \union \Sigma$ is
consistent, contradicting our assumption.

Since $\Gamma \union \Sigma$
is consistent, there is a world $w^*$ in the canonical model such that
$(M^c,w^*) \sat \Gamma \union \Sigma$.  Clearly $\Gamma =
\obji(M^c,w^*)$.  Moreover, by construction, we must have $w^* \in
\K_i^c(w)$.  Thus, $\Gamma \in \Obji(M^c,w)$, as desired.
\eprf

Since there are clearly sets of $i$-sets that are not limit closed,
it follows that this semantics does not satisfy the third property.
One consequence of this
is a result already proved in \cite{Lakemeyer93},
which we reprove
here, using an approach that will be useful for later results.
\pro
\label{not-O-$i$-not-O-j-p}
{\rm \cite{Lakemeyer93}}
If $p \in \Phi$ and $i \ne j$, then $\mdcm \lnot O_i\lnot O_j p$.
\epro

\prf
We proceed by contradiction.  Our goal is to show that if
$(M^c,w) \sat O_i \neg O_j p$, then the set $\Obji(M^c,w)$
must contain a set that includes $O_j p$, for otherwise it would not be
limit closed.  It follows that there is some world $v \in \K_i^c(w)$
such that $(M^c,v) \sat O_j p$, contradicting the assumption that
$(M^c,w) \sat L_i \neg O_j p$.

We first need a definition and a lemma.  We say that a
basic formula $\psi$ is {\em (K45)-independent\/} of
a basic formula $\phi$ if neither $\vdash_{\sWax} \phi \rimp \psi$ nor
$\vdash_{\sWax} \phi \rimp \neg \psi$ hold.

\lem\label{indep} If $n \ge 2$ (\ie there are at least two agents) and
$\phi_1, \ldots, \phi_m$ are consistent basic
$i$-objective formulas, then
there exists a basic $i$-objective formula $\psi$ of the form $L_j\psi'$
which is independent of each of $\phi_1, \ldots, \phi_m$.
\elem

\prf Define the {\em depth\/} of a basic formula $\phi$, denoted
$d(\phi)$, inductively:
\begin{itemize}
\item $d(p) = 0$ for a primitive proposition $p$,
\item $d(\phi \land \psi) = \max(d(\phi), d(\psi))$,
\item $d(\neg \phi) = d(\phi)$,
\item $d(L_i \phi) = 1 + d(\phi)$.
\end{itemize}
Suppose that $\phi_1, \ldots, \phi_m$ are $i$-objective formulas such
that $\max(d(\phi_1(, \ldots, d(\phi_k)) = K$.  Let $p$ be an arbitrary
primitive proposition, and suppose $j \ne i$.  (Such a $j$ exists, since
we are assuming $n \ge 2$.)  Let $\psi$ be the formula $(L_j
L_i)^{K+1} p$, where by $(L_j L_i)^{K+1}$ we mean $K+1$ occurrences of
$L_j L_i$.  Standard model theoretic arguments show that $\psi$ is
independent of $\phi_1, \ldots, \phi_m$.  Very briefly:
By results
of \cite{HM2}, we know that $\phi_i$ is satisfiable in a treelike
structure of depth at most $d(\phi_i)$, for $i = 1, \ldots, m$.  It is
easy to see that this can
be extended to two structures, one of which satisfies $\psi$, and the
other of which satisfies $\neg \psi$.  Hence, $\psi$ is independent of
$\phi_i$.\eprf

Continuing with the proof of Proposition~\ref{not-O-$i$-not-O-j-p},
suppose by way of contradiction that
$(M^c,w) \sat O_i \neg O_j p$.
Let $\overline{w}$ be a world such that $(M^c,\overline{w})\md O_j p$
and let
$\Gamma = \obji(M^c,\overline{w})$.
We claim that $\Gamma$ is a limit of
$\Obji(M^c,w)$.
To see this, consider any finite subset $\Delta$ of $\Gamma$.
Let $\psi$ be an $i$-objective basic formula of the form $L_j
\psi'$ which is independent both of the
conjunction of the formulas in $\Delta$ and of $p$.  The existence of
such a formula follows from Lemma~\ref{indep}.
Let $\Sigma = \subji(M^c,w)$.
By Lemma~\ref{consistency},
$\Sigma \union \Delta \union \{L_j \psi'\}$ is consistent.  Thus,
there is some world $w' \in W^c$
such that $(M^c,w')\sat \Sigma \union \Delta \union \{L_j
\psi'\}$.    By Lemma~\ref{consistency} again, there is some
world $w''$ satisfying $p \land \neg L_j \psi'$ such that $w'' \approx_i
w'$.
Since $(M^c,w') \sat L_j \psi'$, we cannot have $w'' \in \K_j^c(w')$.  It
follows that $(M^c,w') \sat \neg N_j \neg p$, and hence $(M^c,w') \sat
\neg O_j p$.  Moreover, since $w'$ and $w$ agree on all $i$-subjective
formulas, the canonical model construction guarantees that $w' \approx_i
w$.
Since $(M^c,w) \sat O_i \neg O_j p$ , $(M^c,w') \sat \neg O_j p$, and
$w' \approx_i w$, we must have that
$w' \in \K_i^c(w)$.  Thus, $\obji(M^c,w')
\in \Obji(M^c,w)$.  Moreover, by construction,  $\Delta
\subseteq \obji(M^c,w')$.
Since $\Delta$ was chosen arbitrarily, it
follows that $\Gamma$ is a limit of $\Obji(M^c,w)$.  By Lemma~\ref{lc},
$\Gamma \in
\Obji(M^c,w)$.  Thus, there is some world $v \in \K_i^c(w)$ such that
$\obji(M^c,v) = \Gamma$.  It is a simple
property of the canonical-model approach that two worlds that agree on
all basic beliefs of an agent also agree on what the agent only believes.
Hence, since $\overline{w}$ and $v$ agree on $j$'s basic beliefs, it
follows that
$(M^c,v)\md O_j p$, contradicting the assumption that $(\Mc,w) \md
O_i\lnot O_j p$.
\eprf

It may seem unreasonable that $\neg O_i \neg O_j p$ should be valid in
the canonical-model approach.  Why should it be impossible for $i$ to
know only that $j$ does not only know $p$?  After all, $j$ can
(truthfully) tell $i$ that it is not the case that all he ($j$) knows is
$p$.  We return to this issue in Sections~\ref{tree-approach}
and~\ref{synthesis}.  For now, we focus on a proof theory for this
semantics.

\subsection{A Proof Theory}\label{prooftheory}
%
We now consider an axiomatization for the language.
The following axiomatization is exactly like Levesque's except that
axiom {\bf A5} now requires
$\Wax$-consistency
instead of merely
propositional
consistency.
For ease of exposition, we use the same names for the axioms as we did
in the single-agent case with a subscript $n$ to emphasize that we
are looking at the multi-agent version.
\eject
\begin{description}
\item [Axioms:]\ \\[2ex]
\begin{tabular}{ll}
  {\bf A1$_n$}. & Axioms of propositional logic.\\
  {\bf A2$_n$}.
  & $L_i(\alpha \rimp \beta) \rimp (L_i\alpha \rimp L_i\beta)$.\\
  {\bf A3$_n$}.
  & $N_i(\alpha \rimp \beta) \rimp (N_i\alpha \rimp N_i\beta)$.\\
  {\bf A4$_n$}. & $\sigma \rimp L_i\sigma\land N_i\sigma$
              if $\sigma$ is an $i$-subjective formula.\\
  {\bf A5$_n$}. & $N_i\alpha \rimp \lnot L_i\alpha$
             if  $\neg \alpha$ is a $\Wax$-%
consistent
$i$-objective basic formula.
\end{tabular}
\item [Inference Rules:]\ \\[2ex]
\begin{tabular}{lll}
 {\bf MP$_n$.} & From $\alpha$ and $\alpha\rimp\beta$ infer $\beta$.\\
 {\bf Nec$_n$.}
 & From $\alpha$ infer $ L_i\alpha$ and $ N_i\alpha$.\\[1ex]
\end{tabular}
\end{description}
%
Notice that {\bf A5$_n$} assumes that $\alpha$ ranges only over {\em
basic\/} $i$-objective formulas.
We need this restriction in order to appeal to satisfiability in the
existing
logic $\Wax$.\footnote{Note that this peculiar axiom schema is
recursive
since satisfiability in propositional \KFFn\ is decidable~\cite{HM2}.}
To get a more general version of {\bf A5$_n$}, that applies to arbitrary
formulas, we will need to appeal to consistency within the logic that the axioms
are meant to characterize.  We return to this issue in Section~\ref{synthesis}.
%
%
It is not hard to show that these axioms are sound.

\thm
\label{soundness} {\rm \cite{Lakemeyer93}}
For all $\alpha$ in $\onln$, if $\vdash\!\alpha$  then $\mdcm\alpha$.
\ethm

\prf The proof proceeds by the usual induction on the length of a derivation.
%
Here we show only the soundness of {\bf A5$_n$}.
Suppose $\alpha$ is a
basic $i$-objective formula
such that $\neg \alpha$ is $\Wax$-consistent.
Thus, there is an $i$-set containing $\neg \alpha$.  By
Theorem~\ref{all-possibilities-covered}, it follows that for each world
$w \in W^c$,
there is a world $w' \approx_i w$ such that $(M^c,w') \sat \neg \alpha$.
If $w' \in \K_i^c(w)$, it follows that $(M^c,w) \sat \neg L_i \alpha$.
If $w' \notin \K_i^c(w)$, then $(M^c,w) \sat \neg N_i \alpha$.
Thus, $(M^c,w) \sat \neg L_i \alpha \lor \neg N_i \alpha$; equivalently,
$(M^c,w) \sat N_i \alpha \rimp \neg L_i \alpha$.  It follows that
$\sat N_i \alpha \rimp \neg L_i \alpha$.
\eprf

We show in Section~\ref{tree-approach} that this axiomatization is
incomplete.  In fact,
the
formula $\neg O_i \neg O_j p$ is not provable.
Intuitively, part of the problem
here is that
{\bf A5$_n$} is restricted to basic formulas.
For completeness, we would need an analogue of {\bf A5$_n$} for arbitrary
formulas.
\deleted{
Of course, if we could reduce
arbitrary formulas of the form $N_i\beta$ or $L_i\beta$ to
equivalent formulas $N_i\beta^*$ and $L_i\beta^*$, where $\beta^*$
is a basic formula, then we could apply {\bf A5$_n$} directly.
Such a reduction cannot be done
in general. For example, $L_i N_j p$ and $N_i L_j N_i p$ are not
reducible for distinct $i$ and $j$.
However,
if we rule out such cases, the
proof theory is indeed complete for the restricted language, which we call
$\onlnminus$.
}
However, we obtain completeness for a restricted language, which we call
$\onlnminus$.
Roughly, the restriction amounts to limiting what
an agent can only know. In particular, only knowing can only be
applied to basic
formulas.
\dfn
$\onlnminus$ consists of all formulas $\alpha$ in $\onln$ such that, in
$\alpha$, no $N_j$ may occur within the
scope of an $N_i$ or $L_i$ for $i\ne j$.
\edfn
For example, $ N_i L_i\lnot N_i p$ and $ N_i( L_j p\lor N_i\lnot p)$ are
in $\onlnminus$, and $ N_i N_j p$ and $ N_i L_j N_i p$ are not, for
distinct $i$ and $j$.
Note that formulas such as
$O_i\lnot O_j p$ cannot be expressed in \onlnminus.

To prove  completeness for the sublanguage $\onlnminus$,
we need
four preliminary lemmas.  The first describes a normal form for
formulas.  Having such a normal form greatly simplifies the completeness
proof.

\lem\label{normalform} Every formula $\alpha$ in $\onln$ is
provably equivalent to a disjunction of formulas of the following form:
$$\begin{array}{l}
\sigma \land L_1 \phi_{10} \land \neg L_1 \phi_{11}  \land
\ldots \land \neg L_1 \phi_{1m_1} \land \ldots \land L_n \phi_{n0} \land
\neg L_n \phi_{n1} \land \ldots \land \neg L_n \phi_{nm_n} \land\\
N_1 \psi_{10} \land \neg N_1 \psi_{11} \land \ldots
\land \neg N_1 \psi_{1k_1} \land \ldots \land N_n \psi_{n0} \land
\neg N_n \psi_{n1} \land \ldots \land \neg N_n \psi_{nk_n},\end{array}$$
where $\sigma$ is a propositional formula and
$\phi_{ij}$ and $\psi_{ij}$
are all $i$-objective formulas.  Moreover, if $\alpha$ in $\onlnminus$,
we can assume that $\phi_{ij}$ and $\psi_{ij}$ are $i$-objective basic
formulas.
\elem
\prf We proceed by induction on the structure of $\phi$.  The only
nontrivial cases are if $\phi$ is of the form $L_i \phi'$ or $N_i
\phi'$.  If $\phi$ is of the form $L_i \phi'$, then, since $\phi \in
\onlnminus$, $N_j$ does not appear in $\phi'$ for $j \ne i$.  We use the
inductive hypothesis to get $\phi'$ into the normal form described in
the lemma.  Notice that $N_j$ does not appear in the normal form for $j
\ne i$.  We now use the the following equivalences to get $L_i \phi'$
into the normal form:
\begin{itemize}
\item $L_i (\psi \land \psi') \dimp (L_i \psi \land L_i \psi')$
\item $L_i (\psi \lor L_i \psi') \dimp (L_i \psi \lor L_i \psi')$
\item $L_i (\psi \lor \neg L_i \psi') \dimp (L_i \psi \lor \neg L_i
\psi')$
\item $L_i L_i \psi \dimp L_i \psi$
\item $(\neg L_i \false \land L_i \neg L_i \psi) \dimp \neg L_i \psi$
\item $L_i N_i \psi \dimp N_i \psi$
\item $L_i \neg N_i \psi \dimp \neg N_i \psi$.
\end{itemize}
The first five of these equivalences are standard $\Wax$ properties;
the last two are instances of axiom {\bf A4$_n$}.  Similar arguments
work in the case that $\phi$ is of the form
$N_i \phi'$.  We leave the straightforward details to the reader. \eprf

The next two lemmas give us some basic facts about the satisfiability
and validity of formulas in $\OLNm$.

\lem\label{construction1} If $S_j$ is a set of consistent $j$-sets,
$j = 1, \ldots, n$, and
$\sigma$ is a consistent propositional formula, then there is a
$\Wax$ situation $(M,w)$ such that $(M,w) \sat \sigma$ and $\Objj(M,w) =
S_j$. \elem

\prf This follows immediately
from part (b) of Proposition~\ref{construction} below. \eprf

\lem\label{LN} If $\phi$ and $\psi$ are $i$-objective basic formulas
such that
$L_i \phi \land N_i \psi$ is consistent, then $\phi \lor \psi$ is valid.
\elem

\prf Suppose that $\neg \phi \land \neg \psi$ is consistent.  Then,
by Axiom {\bf A5$_n$},
$ N_i (\phi \lor \psi) \rimp \neg L_i (\phi \lor
\psi)$ is provable.  It follows that $N_i \psi \rimp \neg L_i \phi$ is
provable, contradicting the consistency of $L_i \phi \land N_i \psi$.
\eprf

The last lemma gives us a sharper condition on when $w' \notin
\K_i^c(w)$.  This condition will prove useful in the completeness proof.

\lem\label{objOK} If $w, w'$ are worlds in the canonical model such
that $w \approx_i w'$ and
$w' \notin \K_i^c(w)$, then there is an $i$-objective
basic formula $\phi$ such that $L_i \phi \in w$ and $\phi \notin w'$. \elem

\prf By the construction of the canonical model, we know that if
$w' \notin \K_i^c(w)$, then there is some basic formula $\phi$
such that $L_i \phi \in w$ and $\phi \notin w'$.  From
Lemma~\ref{normalform},
it follows that we can assume without loss of generality that
$\phi$ is in the normal form described by that lemma.  Let $A$
consist of all subformulas of $\phi$ that are of the form $L_i \psi$
and do not appear in the scope of any other modal operator.
Let $\phi_A$ be $\band_{\psi \in A \inter w} \psi \land \band_{\psi \in A
- w} \neg \psi$.  It is easy to see that $\phi_A \in w$ (since each of
its conjuncts is).  Since $w \approx_i w'$, it follows that
$\phi_A \in w'$.  Let $\phi'$ be the result of replacing each
subformula $L_i \psi$ of $\phi$ that is in $A$ by either $\true$ or
$\false$, depending on whether $L_i \psi \in w$.  By construction,
$\phi'$ is $i$-objective.
It is easy to
see that $\phi_A \rimp (\phi \dimp \phi')$ is provable.  It follows
that $\phi' \notin w'$.  It also follows that $L_i \phi_A \rimp
(L_i \phi \dimp L_i \phi')$ is provable.  Since $\phi_A$ is an
$i$-subjective formula, $\phi_A \rimp L_i \phi_A$ is provable.  Hence,
$L_i \phi_A \in w$.  Since $L_i\phi \in w$, it follows that $L_i \phi'
\in w$.  This gives us the desired result.  \eprf

We are finally ready to prove the completeness result.

\thm 
\label{completeness} {\rm \cite{Lakemeyer93}}
For all $\alpha\in\onlnminus$, if $\mdcm\alpha$ then $\vdash\!\alpha$.
\ethm
\prf
As usual, it suffices to show that if the formula $\alpha \in \onlnminus$
is consistent, then it is satisfiable in the canonical model.
Without loss of generality, we can assume that $\alpha$ is in the normal
form described in Lemma~\ref{normalform}:
$$\begin{array}{l}
\sigma \land L_1 \phi_{10} \land \neg L_1 \phi_{11}  \land
\ldots \land \neg L_1 \phi_{1m_1} \land \ldots \land L_n \phi_{n0} \land
\neg L_n \phi_{n1} \land \ldots \land \neg L_n \phi_{nm_n} \land\\
N_1 \psi_{10} \land \neg N_1 \psi_{11} \land \ldots
\land \neg N_1 \psi_{1k_1} \land \ldots \land N_n \psi_{n0} \land
\neg N_n \psi_{n1} \land \ldots \land \neg N_n \psi_{nk_n}.\end{array}$$
Moreover, since $\alpha \in
\onlnminus$, we can assume that $\phi_{ij}$ and $\psi_{ij}$ are
$i$-objective basic formulas.  Let $A_i$ consist of all the consistent
formulas of the form $\phi_{i0} \land \psi_{i0} \land \neg \phi_{ij}$
or $\phi_{i0} \land \psi_{i0} \land \neg \psi_{ij}$, $j \ge 1$.
Let $\xi_i$ be a formula that is independent of all the formulas in
$A_i$; such a formula exists by Lemma~\ref{indep}.  Let $S_i$ consist of
all $i$-sets containing $\phi_{i0} \land (\neg \psi_{i0} \lor
(\psi_{i0} \land \xi_i))$.
By Lemma~\ref{construction1}, there is a $\Wax$ structure
$(M,w)$ such that $\Obji(M,w) = S_i$, $i = 1, \ldots, n$, and $(M,w)
\sat \sigma$.
Thus, there must be a world $w^*$ in the canonical model such that
$w^* = \{\mbox{basic } \phi' \;|\; (M,w) \sat \phi'\}$.  We claim that
$(M^c,w^*) \sat \alpha$.

To see this, let $\alpha'$ be the formula
$\sigma \land L_1 \phi_{10} \land \neg L_1 \phi_{11}  \land
\ldots \land \neg L_1 \phi_{1m_1} \land \ldots \land L_n \phi_{n0} \land
\neg L_n \phi_{n1} \land \ldots \land \neg L_n \phi_{nm_n}$.  We first
show that
$(M,w) \sat \alpha'$.
By construction, we have that $(M,w) \sat \sigma$.
Furthermore,
by definition, each world $w' \in \K_i^M(w)$ satisfies $\phi_{i0}$,
so we have that $(M,w) \sat L_i \phi_{i0}$.
Since $L_i \phi_{i0} \land \neg L_i \phi_{ij}$ is consistent for each $j
\ge 1$, it must be the case that $\phi_{i0} \land \neg \phi_{ij}$ is
consistent.  Thus, one of $\phi_{i0} \land \neg \psi_{i0} \land \neg
\phi_{ij}$ or $\phi_{i0} \land \psi_{i0} \land \neg \phi_{ij}$ is
consistent.  If the latter is consistent, then by the choice of $\xi$,
$\phi_{i0} \land \psi_{i0} \land \xi_i \land \neg \phi_{ij}$ must be
consistent as well.  Since $S_i$ consists of all $i$-sets containing
$\phi_{i0} \land (\neg \psi_{i0} \lor
(\psi_{i0} \land \xi_i))$,
it follows that there must be an
$i$-set in $S_i$ containing $\neg \phi_{ij}$.
It follows that $(M,w) \sat \neg L_i \phi_{ij}$, for $j \ge 1$.
Thus, we have shown that $(M,w) \sat \alpha'$.
Since
$(M,w)$ and $(M^c,w^*)$ agree on basic formulas, it follows that
$(M^c,w^*) \sat \alpha'$.

Next, we show that $(M^c,w^*) \sat
N_1 \psi_{10} \land \ldots \land N_n \psi_{n0}$.  To this end, suppose
that
$w' \approx_i w^*$ and $w' \notin \K_i^c(w^*)$.  By Lemma~\ref{objOK},
there must be
some $i$-objective basic formula $\phi'$
such that $L_i \phi' \in w^*$
and $\neg \phi' \in w'$.  Since $L_i \phi' \in w^*$, it follows that
$(M,w) \sat L_i \phi'$, and hence $\phi'$ is in every $i$-set in $S_i$.
It follows that $\obji(w') \notin S_i$.  Now, one of the following
four formulas must be in $\obji(w')$: (1) $\phi_{i0} \land \psi_{i0}$,
(2) $\phi_{i0} \land \neg \psi_{i0}$,
(3) $\neg \phi_{i0} \land \psi_{i0}$, (4)  $\neg \phi_{i0} \land
\neg \psi_{i0}$.  Since $L_i \phi_{i0} \land N_i \psi_{i0}$ is
consistent, it cannot be (4), by Lemma~\ref{LN}.  It cannot be (2), for
otherwise $w'$ would be in $S_i$.  Thus, it must be (1) or (3), so
$\psi_{i0} \in \obji(w')$.  Since this is true for all $w'$ such
that $w' \approx_i w^*$ and $w' \notin \K_i^c(w^*)$, it
follows that $(M^c,w^*) \sat N_i \psi_{i0}$, for $i = 1, \ldots, n$.

Finally, we must show that $(M^*,w^*) \sat \neg N_1 \psi_{ij}$, for $i =
1, \ldots, n$ and $j = 1, \ldots, k_i$.  Clearly $\psi_{i0} \land \neg
\psi_{ij}$ is consistent, for otherwise $N_i \psi_{i0} \land \neg N_i
\psi_{ij}$ would be inconsistent.  Thus, at least one of
(1) $\psi_{i0} \land \neg \psi_{ij} \land \neg \phi_{i0}$
or (2) $\psi_{i0} \land \neg \psi_{ij} \land \phi_{i0}$ is consistent.
In case (2), by choice of $\xi_i$, the formula $\psi_{i0} \land \neg
\psi_{ij} \land \phi_{i0} \land \neg \xi_i$ is consistent.  Let $\beta$
be $\psi_{i0} \land \neg \psi_{ij} \land \neg \phi_{i0}$ if it is
consistent, and $\psi_{i0} \land \neg
\psi_{ij} \land \phi_{i0} \land \neg \xi_i$ otherwise.  By construction,
$\beta$ is consistent.
 By Lemma~\ref{consistency}, there is a situation
$(M',v)$ such that $\subji(M',v) = \subji(M^*,w^*)$ and $(M',v)
\sat \beta$.  There is a world $w''$ in the canonical model which
agrees with $(M',v)$ on the basic formulas.  By construction, we have
$w'' \approx_i w^*$.  Moreover, since $(M^c,w) \sat L_i(\phi_{i0} \land
(\neg \psi_{i0} \lor (\psi_{i0} \land \xi)))$, it follows that
$(M^c,w^*)
\sat L_i \neg \beta$.  Since $(M^c,w') \sat \beta$, we have that
$w' \notin \K_i^c(w^*)$.  Moreover, since $(M^c,w') \sat \neg
\psi_{ij}$, it follows that $(M^c,w^*) \sat \neg N_i \psi_{ij}$, as
desired.  This completes the proof. \eprf





\subsection{Discussion}
As we have shown,
the canonical-model semantics for $N_i$ has some attractive features,
in particular when restricted to the language $\onlnminus$.  It is for
this sublanguage that we have
a nice proof-theoretic characterization.
%
There is some evidence, however,
that
the semantics may not have the behavior we desire when we move beyond
$\onlnminus$.  For one thing,
the formula $\neg O_i \neg O_j p$ is valid in the canonical model:~it
is impossible that all $i$ knows is that it is not the case that all $j$
knows is $p$.  While it is certainly consistent for $\neg O_i \neg O_j
p$ to hold, it seems reasonable to have a semantics that allows $O_i
\neg O_j p$
to be hold as well.  As we have seen, the validity of $\neg O_i \neg O_j
p$ follows from the fact that the canonical-model
semantics does not have the third property of Levesque's semantics in
the single-agent case: not all subsets of conceivable states are
possible.  In the next section, we discuss a different
approach to giving
semantics to only knowing---essentially that taken in \cite{Hal11}---%
that has all three of Levesque's properties,
at least as long as we continue to represent an agent's objective
state of affairs using basic formulas only.
The approach agrees with the canonical-model approach on formulas in $\onlnminus$,
but makes $O_i \neg O_j p$ satisfiable.  Unfortunately, as we shall
see, it too suffers from problems.

\section{The $i$-Set Approach}\label{tree-approach}

In the $i$-set approach, we maintain the intuition that
the set of conceivable states for each agent $i$ can be
identified with  the set of $i$-sets.  We no longer restrict attention
to the canonical model though; we consider all Kripke structures.

We define a new semantics
$\sat'$ as follows:  all the clauses of $\sat'$ are identical to the
corresponding clauses for $\sat$, except that for $N_i$.  In this case,
we have
$$\begin{array}{l}
(M,w) \sat' N_i \phi \mbox{ iff } (M',w') \sat' \phi \mbox{ for all
situations $(M',w')$ such that}\\
\ \ \ \ \mbox{$\Obji(M,w) = \Obji(M',w')$ and
$\obji(M',w') \notin \Obji(M,w)$.}\end{array}$$
Notice that $\sat$ and $\sat'$ agree for basic formulas; in general, as
we shall see, they differ.  We remark that this definition is equivalent
to the one given
in \cite{Hal11}, except that there, rather than $i$-sets, {\em
$i$-objective
trees\/} were considered.  We did not want to go through the overhead
of introducing $i$-objective trees here, since it follows from results
in \cite{Hal11,Hal13}
that $i$-sets are equivalent to $i$-objective trees: every $i$-set
uniquely determines an $i$-objective tree and vice versa.

How well does this approach fare in terms of computing the truth of a
formula at a given world.  Since we now allow arbitrary
structures,
not just the canonical model, it seems that for computational reasons,
it seems we ought to focus on finite structures.  However, when it comes
to formulas of the form $N_i \alpha$, where $\alpha$ is $i$-objective,
satisfiable formula, it can be shown that $(M,w) \sat' \neg N_i \alpha$.
There are simply too few sets in $\Obji(M,w)$ if $M$ is finite to
have $\neg \alpha$ hold in all ``impossible'' situations.  This means
that to really model interesting situations with only knowing, we must
use infinite models, and we are back to the problems discussed in the
case of the canonical model.  Thus, again we focus on whether this
approach gives reasonable semantics to the $N_i$ operator.

Notice that to decide if $N_i
\phi$ holds in $(M,w)$, we consider all
situations that agree with $(M,w)$ on the set of possible states, hence
this semantics satisfies the first of the three properties we
isolated
in the single-agent case.  It is also clear that the $i$-sets considered
in evaluating the truth of $N_i \phi$ are precisely those not considered
in evaluating the truth of $L_i \phi$; hence we satisfy the second
property.  Finally, as we now show, for every set $S$ of $i$-sets,
there is a situation $(M,w)$ such that $\Obji(M,w) = S$.
%
In fact,
we prove an even stronger result.
\dfn
Let \newl{$\objsi(M,w)$} consist of all $i$-objective formulas (not necessarily
just $i$-objective basic formulas) true at $(M,w)$ (with respect to
$\sat'$) and let \newl{$\Objsi(M,w)$} =
$\{\objsi(M,w') \;|\; w' \in \K_i^M(w)\}$.
\edfn

\pro\label{construction}
Let $\Gamma$ be a satisfiable set of $i$-objective formulas, let $S_i$
be a set of maximal satisfiable sets of $i$-objective
formulas, $i = 1, \ldots, n$, let $\Sigma$ be a satisfiable set of
$i$-subjective formulas, and let $\sigma$ be a satisfiable propositional formula.
Then
\begin{enumerate}
\item[(a)] there exists a situation $(M_1,w_1)$ such that $\Gamma
\subseteq \objsi(M_1,w_1)$ and $S_i = \Objsi(M_1,w_1)$.
\item[(b)] there exists a situation $(M_2,w_2)$ such that $(M_2,w_2)
\sat \sigma$ and $\Objsj(M_2,w_2) = S_j$, $j = 1, \ldots, n$.
\item[(c)] there exists a situation $(M_3,w_3)$ such that $(M_3,w_3)
\sat \Gamma \land \Sigma$.
\end{enumerate}
\epro

\prf For part (a), we first show that, given an arbitrary situation
$(M,w)$, we can construct a situation $(M^*,w^*)$ such that
$\objsi(M^*,w^*) = \objsi(M,w)$ and there are no worlds $i$-accessible
from $w^*$.  The idea is to have $M^*$ be the result of adding $w^*$
to the worlds in $M$, where $w^*$ is just like $w$
except that it has no $i$-accessible worlds and $w^*$ is not accessible
from any world.
More formally, if $M = (W,\pi,\K_1, \ldots,\K_n)$, we take $M^* =
(W^*,\pi^*,\K_1^*,\ldots, \K_n^*)$, where $W^* = W \union \{w^*\}$,
$\pi^*(w')
= \pi(w')$ for $w' \in W$, $\pi^*(w^*) = \pi(w)$, $\K_j^* = \K_j \union
\{(w^*,w') \;|\; w' \in \K_j(w)\}$ for $j \ne i$,
and $\K_i^* = \K_i$.  It is easy to see that $\K_j^*$ is Euclidean and
transitive.  By construction, there are no worlds $i$-accessible from
$w^*$ and $(w',w^*) \notin \K_j^*$ for all $w'$ and all $j$.
Moreover, if $\psi$ is an $i$-objective formula,
we have $(M^*,w^*) \sat' \psi$ iff $(M,w) \sat' \psi$, since for $j \ne
i$, we have $\K_j(w^*) = \K_j(w)$.  In particular, this means that
$\objsi(M^*,w^*) = \objsi(M,w)$.

For each $\Delta \in S' =
S_i \union \{\Gamma\}$, there is a situation
$(M^\Delta,w^\Delta) \sat' \Delta$, where $M^\Delta = (W^\Delta,
\pi^\Delta, \K_1^\Delta, \ldots, \K_n^\Delta)$. By the argument above, we
can assume without loss of generality that there are no worlds
$i$-accessible from $w^\Delta$ and $w^\Delta$ is not accessible from
any world.
We define $M_1 = (W,\pi,\K_1, \ldots,
\K_n)$ by taking $W$ to be the union of all the worlds in $W^\Delta$,
$\Delta \in S'$.  (We can assume without
loss of generality that these are disjoint sets of worlds.)
We define $\pi$ so that $\pi|_{W^\Delta} = \pi^\Delta$.  We define
$\K_j = \union_{\Delta \in S'} \K_j^\Delta$ for $j \ne i$, and
$\K_i$ to be the least transitive, Euclidean set containing
$\union_{\Delta \in S'} \K_i^\Delta \union \{(w^\Gamma,
w^\Delta) \;|\; \Delta \in S\}$.  It is easy to check that
$\objsi(M_1,w^\Delta) = \objsi(M^\Delta,w^\Delta)$ (although this
depends on the fact that $w_\Delta$ is not $j$-accessible
from any world for $j \ne i$).  Thus, $\Objsi(M_1,w^\Gamma) = S_i$
and $\objsi(M_1,w^\Gamma) \supseteq \Gamma$.  Thus, we can take
$w_1 = w^\Gamma$, completing the proof of part (a).

To summarize the construction of part (a), we start with an arbitrary
situation $(M,w)$ satisfying $\Gamma$, convert it to a
situation satisfying $L_i \false \land \Gamma$, essentially by
modifying the $i$-accessibility relation at $w$ so that there are no
worlds $i$-accessible from $w$ and $w$ is not accessible from any world,
and then again modifying the
$i$-accessibility relation at $w$ so that we get a structure $(M_1,w_1)$
such that $\Objsi(M_i,w_i) = S_i$.  Note that in doing this
construction, we did not change the propositional formulas true at $w$,
nor did we change the worlds that were $j$-accessible from $w$ for
$j \ne i$.  Thus, starting with a situation that satisfies a
propositional formula $\sigma$, we can repeat this construction for each
$i$ in turn, for $i = 1, \ldots, n$.  The resulting situation is
$(M_2,w_2)$, and it clearly has the desired properties.
This proves part (b).

For part (c), suppose $(M',w') \sat \Sigma$; let $\Objsi(M',w') =
S_i$.  By part (a), there is a situation $(M_3,w_3)$ such that
$\Objsi(M_3,w_3) = S_i$ and $(M_3,w_3) \sat \Gamma$.  Since the
set of subjective formulas true at a situation $(M,w)$ is completely
determined by $\Objsi(M,w)$, and $\Objsi(M',w') = \Objsi(M_3,w_3)$,
it follows that $(M_3,w_3) \sat \Sigma$ as well.  \eprf


How does this semantics compare to the canonical model semantics?
First of all, it is easy to see that the axioms are sound.
We write $\sat' \phi$ if $(M,w) \sat' \phi$ for every situation $(M,w)$.
Then we have the following result.
\thm {\rm \cite{Hal11}}
For all $\alpha \in \onln$, if $\vdash \alpha$ then $\sat' \alpha$.
\ethm

\prf As usual, the proof is by induction on the length of a
derivation.  All that needs to be done is to show that all the axioms
are sound.  Again, this is straightforward.  The proof in the case of
{\bf A5$_n$} proceeds just as that in the proof of
Theorem~\ref{soundness}, using the fact that this semantics satisfies
Levesque's second property. \eprf

Moreover, we again get completeness for the sublanguage $\onlnminus$.

\thm {\rm \cite{Hal11}}
For all $\alpha \in \onlnminus$,
$\vdash \alpha$ iff $\sat'
\alpha$. \ethm

\prf As usual, it suffices to show that if $\alpha$ is consistent
with the axioms, then $\alpha$ is satisfiable under the $\sat'$
semantics.  From Theorem~\ref{completeness}, we know that
$\alpha$ is satisfiable in the canonical model under the $\sat$
semantics.   Thus, it suffices to show that for all formulas $\alpha
\in \onlnminus$, we have $(M^c,w) \sat \alpha$ iff $(M^c,w) \sat'
\alpha$.
By Lemma~\ref{normalform}, it suffices to consider formulas
$\alpha$ in normal form.  We proceed by induction on the structure of
formulas.  The only nontrivial case obtains if $\alpha$ is of the form $N_i
\alpha'$.  Since $\alpha$ is in normal form, we can assume that
$\alpha'$ is basic.  Suppose $(M^c,w) \sat' N_i \alpha'$.  To show that
$(M^c,w) \sat N_i \alpha'$, we must show that if $w' \approx_i w$ and
$w' \notin \K_i^c(w)$, then $(M^c,w') \sat \alpha$.  By definition,
if $w'
\approx_i w$, then $\Obji(M^c,w') = \Obji(M^c,w)$.  Moreover, we must
have $\obji(M^c,w') \notin \Obji(M^c,w)$, for otherwise we would have
$w' \in \K_i^c(w)$.  Hence, we must have $(M^c,w') \sat' \alpha'$.  By
the induction hypothesis, we have $(M^c,w') \sat \alpha'$.  Thus,
$(M^c,w) \sat N_i \alpha'$, as desired.

For the converse, suppose that
$(M^c,w) \sat N_i \alpha'$.  We want to show that $(M^c,w) \sat' N_i
\alpha'$.  Suppose that $(M',w')$ is such that $\Obji(M',w') =
\Obji(M^c,w)$ and $\obji(M',w') \notin \Obji(M^c,w)$.
We must show that $(M',w') \sat' \alpha$.
It is easy to
see that for every
situation $(M,w)$ and basic formula $\phi$, we have that $(M,w) \sat L_i
\phi$ iff $(M,w) \sat' L_i \phi$ iff $\phi$ is in every set in
$\Obji(M,w)$.  Thus, it follows that $\subji(M',w') = \subji(M^c,w)$.
There must be a world $w''$ in $M^c$ such that $(M^c,w'')$ agrees
with $(M',w')$ on all basic formulas according to the $\sat$ semantics.
Since $\subji(M^c,w'') = \subji(M^c,w)$, it follows from
Lemma~\ref{$i$-equivalence-same-as-same-basic-beliefs} that $w \approx_i
w''$.  Since $\obji(M',w') \notin \Obji(M^c,w)$ and $\obji(M',w') =
\obji(M^c,w'')$, it follows that $w'' \notin \K_i^c(w)$.  Since $(M^c,w)
\sat N_i \alpha'$, we must have that $(M^c,w'') \sat \alpha'$.  And
since $(M^c,w'')$ and $(M',w')$ agree on basic formulas, it follows that
$(M',w') \sat \alpha'$.  Finally, since $\sat$ and $\sat'$ agree for
basic formulas, we have $(M',w') \sat' \alpha'$.
This completes the proof that $(M,w) \sat N_i \alpha'$. \eprf

Although our axiomatization is complete for $\onlnminus$, as we now
show, it is not complete for the full language,
for neither $\sat$ nor $\sat'$.
Since the axiomatization is sound for both $\sat$ and $\sat'$, to prove
incompleteness, it suffices to provide a formula which is satisfiable
with respect to $\sat'$ and not $\sat$, and another formula
which is satisfiable  with respect to $\sat$ and not $\sat'$.
As is shown in Proposition~\ref{incomplete1},
$O_i \neg O_j p$ is satisfiable
with respect to $\sat'$ and (by Proposition~\ref{not-O-$i$-not-O-j-p})
not with respect to $\sat$.  On the other hand, it is easy to see
that $L_j \false \land N_j \neg O_i \neg O_j p$ is satisfiable
with respect to $\sat$ (in fact, it is equivalent to $L_j \false$);
as shown in Proposition~\ref{incomplete2},
it is not satisfiable with respect
to $\sat'$.


\pro\label{incomplete1} $O_i \neg O_j p$ is
satisfiable under the $\sat'$ semantics. \epro

\prf
Let $S = \{\objsi(M,w) \;|\; (M,w) \sat' \neg O_j p\}$.
By Proposition~\ref{construction}, there is a situation
$(M^*,w^*)$ such that $\Objsi(M^*,w^*) = S$.
We claim that $(M^*,w^*) \sat' O_i \neg O_j p$.  Clearly
$(M^*,w^*) \sat' L_i \neg O_j p$, since $\neg O_j p$ is true at all
worlds $i$-accessible from $w^*$.  To see that $(M^*,w^*) \sat' N_i O_j
p$, suppose that $\Obji(M,w) = \Obji(M^*,w^*)$ and $\obji(M,w)
\notin \Obji(M^*,w^*)$.  We want to show that $(M,w) \sat' O_j p$.
Suppose that $(M,w) \sat' \neg O_j p$.
By definition, $\objsi(M,w) \in S$, so there is some world
$w' \in \K_i^{M^*}(w^*)$ such that $\objsi(M^*,w') =
\objsi(M,w)$.  In particular, this means that $\obji(M^*,w') =
\obji(M,w)$.  But this contradicts the assumption that
$\obji(M,w) \notin \Obji(M^*,w^*)$.  Thus, $(M,w) \sat O_j p$ as
desired, and $(M^*,w^*) \sat O_i \neg O_j p$. \eprf

\pro\label{incomplete2}  There is a formula $\beta$ such that $\sat'
\beta$ but $\neg \beta$ is satisfiable under the $\sat$ semantics.
\epro

\prf
We first show that if $\phi$ is an $i$-objective
formula that is satisfiable under the $\sat'$ semantics, then $\sat' L_i
\false \rimp \neg N_i \neg \phi$.  For suppose that $\phi$ is
satisfiable in a situation $(M,w)$.  By Proposition~\ref{construction},
there is a situation $(M^*,w^*)$ such that $\objsi(M^*,w^*) =
\objsi(M,w)$ and $\Objsi(M^*,w^*) = \emptyset$.  This means that
$(M^*,w^*) \sat' \phi \land L_i \false$.
Now let $(M',w')$ be any situation
satisfying $L_i \false$.  Then $\Obji(M',w') = \Obji(M^*,w^*) =
\emptyset$, and $\obji(M^*,w^*) \notin \Obji(M',w')$.  It follows that
$(M',w') \sat' \neg N_i \neg \phi$.  Thus, we have shown that
$\sat' L_i \false \rimp \neg N_i \neg \phi$.
Since, as we showed
in Proposition~\ref{incomplete1}, the formula $O_j \neg O_i p$ is
satisfiable, this means that
$\sat' L_i \false \rimp \neg N_i \neg O_j \neg O_i p$.
On the other hand, since
$O_j \neg O_i p$ is not satisfiable with respect to $\sat$, as we showed
in Proposition~\ref{not-O-$i$-not-O-j-p},
neither is $\neg N_i \neg O_j \neg O_i p$, and hence $L_i \false \rimp
\neg N_i \neg O_j \neg O_i p$ is not valid under the $\sat$ semantics.
Indeed, $L_i \false \land N_i \neg O_j \neg O_i p$ is equivalent to $L_i
\false$
under the $\sat$ semantics.
\eprf

We can now show that our axiom system is incomplete for the full
language with respect to both the $\sat$ and $\sat'$ semantics.

\thm\label{incompleteness} There exist formulas $\alpha$ and $\beta$
in $\onln$ such that $\not\vdash \alpha$ and $\sat \alpha$,
and $\not\vdash \beta$ and $\sat' \beta$.
\ethm

\prf By Propositions~\ref{not-O-$i$-not-O-j-p}
and~\ref{incomplete1}, we have that $\sat \neg O_j
\neg O_i p$, but
$\not \sat' \neg O_j \neg O_i p$.  Since $\vdash$ is sound with respect
to $\sat'$, we cannot have $\vdash \neg O_j \neg O_i p$ (for otherwise
we would have $\sat' \neg O_j \neg O_i p$).  Thus, we can take $\alpha$
to be $\neg O_j \neg O_i p$.  A similar argument shows we can take
$\beta$ to be $L_i \false \rimp \neg N_i \neg O_j O_i p$.
\eprf

The fact that neither $\sat$ nor $\sat'$ is complete with respect to the
axiomatization described
earlier
is not necessarily
bad.  We may be able to find a natural complete axiomatization.
However, as we suggested above, the fact that $\neg O_j \neg O_i p$ is
valid under the $\sat$ semantics suggests that this semantics does not
quite satisfy our intuitions with regards to only-knowing for formulas
in $\onln - \onlnminus$.  As we now show,
$\sat'$ also has its
problems.
We might hope that if $\phi$ is a satisfiable $i$-objective
formula, then $N_i \phi \rimp \neg L_i \phi$ would be valid under the
$\sat'$ semantics.  Unfortunately, it is not.
\pro\label{problems} The formula
$N_i \neg O_j p \land L_i \neg O_j p$ is satisfiable under the
$\sat'$ semantics.  \epro

\prf
First we show that for any situation $(M,w)$
that satisfies $O_j p$,  there exists another situation
$(M',w')$ such that $(M,w)$ and $(M',w')$ agree on all basic
formulas, but $(M',w') \sat' \neg O_j
p$.  We can construct $(M',w')$ as follows: Choose a particular
set $\Gamma \in \Objj(M,w)$.  It easily follows from
Proposition~\ref{construction} that there is a situation $(M',w')$
such that $\Objj(M',w') = \Objj(M,w) - \{\Gamma\}$ and
$\objj(M',w') = \objj(M,w)$.
We now show that for any
basic formula $\phi$, we have $(M,w) \sat' \phi$ iff $(M',w') \sat'
\phi$.  If $\phi$ is a $j$-objective formula, this is immediate from
the construction.  Thus, it suffices to deal with the case that
$\phi$ is of the form $L_j \phi'$.  By Lemma~\ref{normalform}, we can
assume without loss of generality that $\phi'$ is $j$-objective.
Suppose that $\phi' $ is a consistent $j$-objective formula.
In this case, it is almost immediate from
the definitions that if
$(M,w) \sat' L_j \phi'$ then $(M',w') \sat' L_j \phi'$.
For the converse,
suppose that $(M,w) \sat' \neg L_j \phi'$.  Then there is some world
$w'' \in \K_j^M(w)$ such that $(M,w'') \sat \neg \phi'$.  Since $(M,w)
\sat' L_j p$, we have that $(M,w'') \sat' p \land \neg \phi'$.
Let $\psi$ be a $j$-objective basic formula that is independent of $p
\land \neg \phi'$.   Let $\phi''$ be $\phi' \land p \land \neg
\psi$ if $\psi \in \Gamma$, and $\phi' \land p \land \psi$ otherwise.
Since $\psi$ is independent of $p \land \neg \phi'$, it follows that
$\phi''$ is consistent.  Let $\Delta$ be any $j$-set
containing $\phi''$.  It must be the case that $\Delta
\in \Objj(M,w)$, for if not, let $(M^*,w^*)$ be a situation such that
$\Objj(M^*,w^*) = \Objj(M,w)$ and $\objj(M^*,w^*) = \Delta$ (such a
situation exists by Proposition~\ref{construction}).  Then
$(M^*,w^*) \sat p$, contradicting the assumption that $(M,w) \sat N_j
\neg p$.  By construction, $\Delta \ne \Gamma$.  Thus, there is some
world $v \in \K_j^{M'}(w')$ such that $\objj(M',v) = \Delta$.
It follows that $(M',v) \sat' \neg \phi'$, so
$(M',w') \sat' \neg L_j
\phi'$. Thus, $(M',w')$ agrees
with $(M,w)$ on all basic formulas.  However,
since $\Gamma \notin \Objj(M,w)$, it follows that $(M',w') \sat
\neg N_j \neg p$, and hence that $(M',w') \sat \neg O_j p$.

Let $S = \{\objsi(M,w) \;|\; (M,w) \sat' \neg O_j p\}$.  By
Proposition~\ref{construction},
there is a situation $(M^*,w^*)$ such that $\Objsi(M^*,w^*) = S$.
Clearly  $(M^*,w^*)
\sat' L_i \neg O_j p$.  We now show that $(M^*,w^*) \sat' N_i \neg O_j
p$ as well.  For suppose that $(M,w)$ is a
situation such that $\Obji(M,w) = \Obji(M^*,w^*)$ and
$\obji(M,w) \notin \Obji(M^*,w^*)$.  Moreover, suppose, by way of
contradiction, that $(M,w) \sat' O_j p$.  By the arguments above,
it follows that there is a situation $(M',w')$ such that
$(M',w') \sat' \neg O_j p$ and $(M,w)$ and $(M',w')$ agree on all basic
formulas.   By construction,
$\objsi(M',w') \in S = \Objsi(M^*,w^*)$, so
$\obji(M',w') \in \Obji(M^*,w^*)$.  Since
$\obji(M,w) = \obji(M',w')$,
we must also have $\obji(M,w) \in \Obji(M^*,w^*)$, contradicting the
choice of $(M,w)$.  Thus, $(M,w) \sat \neg O_j p$, as desired, and so
$(M^*,w^*) \sat L_i \neg O_j p \land N_i \neg O_j p$. \eprf

\subsection{Discussion}
Proposition~\ref{problems} shows that
although the $i$-set semantics has the three
properties we claimed were appropriate, $N_i$ and $L_i$ still do
not always interact in what seems to be the appropriate way.
Intuitively, the problem here is
that there is more to $i$'s view of a world than just the $i$-objective
basic formulas that are true there.  We should really identify
$i$'s view of a situation $(M,w)$ with
the set of {\em all\/} $i$-objective formulas that are true there.
In the canonical-model approach,
the $i$-objective basic formulas that are true at a world
can be shown to
determine
all the $i$-objective formulas that are true at that world.
This is not true at all
situations under the $i$-set approach.

Indeed, it is no longer true
that the $i$-set approach has the second of the
three properties once we take $i$'s
view of $(M,w)$ to be $\objsi(M,w)$.  For consider
the situation $(M^*,w^*)$ constructed in the proof of
Proposition~\ref{problems}.
As the proof of that lemma shows, $\{\objsi(M^*,w) \;|\; w \in
\K_i^{M^*}(w^*)\} \union \{\objsi(M',w') \;|\; \obji(M',w') \notin
\Obji(M^*,w^*)\}$ does not include all maximal sets of $i$-objective
formulas.  In particular, it does not include those maximal sets
that satisfy $O_j p$.

To summarize: while the i-set approach arguably has enough worlds,
it suffers from the fact that the full complement of worlds
is not always taken into account for $L_i$ and $N_i$, as the
above example shows. While the canonical model approach
does not run into this problem, it suffers from a perhaps more basic
deficiency: there just are not enough worlds to begin with. For example,
there is no world where $O_i O_j p$ is true.

We consider a different approach in the
next section that attempts to deal with both problems.
\section{What Properties Should Only Knowing Have?}\label{synthesis}
Up to now, we have provided two semantics for only knowing.  While both have
properties we view as desirable, they also have properties that seem somewhat
undesirable.  This leads to an obvious question: What properties should only
knowing have?  Roughly speaking, we would like to have the multi-agent version
of Levesque's axioms, and no more. Of course, the problem here is axiom {\bf
A5$_n$}.  It is not so clear what the multi-agent version of that should be.
The problem is one of circularity: We would like to be able to say that $N_i
\phi \rimp \neg L_i \phi$ should hold for any
consistent
$i$-objective formula.  The problem is that in order to say what
the
consistent
formulas are, we need to define the axiom system.
In particular, we have to make precise what this axiom should be.

To deal with this problem, we extend the language so that we can
explicitly talk about satisfiability and validity in the language.
We add a modal operator $\Vall$ to the language. The formula $\Val{\phi}$
should be read ``$\phi$ is valid''.  Of course, its dual $\Con{\phi}$,
defined as $\neg \Val{\neg \phi}$, should be read ``$\phi$ is
satisfiable''.  With this operator in the language, we can
replace {\bf A5$_n$} with
\begin{itemize}
\item[{\bf A5$'_n$}.]
$\Con{\neg \alpha} \rimp (N_i\alpha \rimp \neg L_i \alpha)$
if $\alpha$ is $i$-objective.
\end{itemize}

In addition, we have the following rules for reasoning about
validity and satisfiability:

\begin{description}
\item[{\bf V1}.] $(\Val{\phi} \land \Val{\phi \rimp \psi}) \rimp
\Val{\psi}$.
\item[{\bf V2}.]
$\Con{\phi}$, if $\phi$ is a satisfiable propositional formula.%
\footnote{We
can replace this by the simpler $\Con{p_1' \land \ldots \land p_k'}$,
where $p_i'$ is a literal---either a primitive proposition or its
negation---and $p_1' \land \ldots \land p_k'$ is
consistent.}
\item[{\bf V3}.]
$(\Con{\alpha \land \beta_1} \land \ldots \land
\Con{\alpha \land \beta_k} \land
\Con{\gamma \land \delta_1} \land \ldots \land
\Con{\gamma \land \delta_m} \land
\Val{\alpha \lor \gamma})  \rimp$\\
$\Con{L_i \alpha \land \neg L_i \neg \beta_1
\land \ldots \land \neg L_i \neg \beta_k \land N_i \gamma \land
\neg N_i \neg \delta_1 \land \ldots \land \neg N_i \neg \delta_m}$,\\
if $\alpha, \beta_1, \ldots, \beta_k,
\gamma, \delta_1, \ldots, \delta_m$ are $i$-objective
formulas.
\item[{\bf V4}.]
$(\Con{\alpha} \land \Con{\beta}) \rimp \Con{\alpha \land \beta}$
if $\alpha$ is $i$-objective and $\beta$ is $i$-subjective.
\item[{\bf Nec$_V$}.] From $\phi$ infer $\Val{\phi}$.
\end{description}
Axiom {\bf V1} and the rule {\bf Nec$_V$} make $\Vall$ what is
called a
{\em normal\/} modal operator.  In fact,
it can be shown to satisfy
all the axioms of S5.
The interesting clauses are clearly
{\bf V2}--{\bf V4}, which capture the intuitive properties of
validity and satisfiability.

If we restrict to basic formulas,
then {\bf V3} simplifies to
$(\Con{\alpha \land \beta_1} \land \ldots \land
\Con{\alpha \land \beta_k}) \rimp
\Con{L_i \alpha \land \neg L_i \neg \beta_1
\land \ldots \land \neg L_i \neg \beta_k}$ (we can take $\gamma,
\delta_1, \ldots, \delta_m$ to be
$\true$ to get this).
The soundness of this
axiom (interpreting $\Conn$ as satisfiability) follows using much
the same arguments as those in the proof of Proposition~\ref{construction}.
The soundness of {\bf V4} if we restrict to basic formulas follows
from Lemma~\ref{consistency}.  More interestingly, it follows from
the completeness proof given below that these axioms completely
characterize satisfiability in $\Wax$; together with the $\Wax$ axioms,
they provide a sound and complete language for the language augmented
with the $\Vall$ operator.

Let AX$'$ consist of the axioms for $\onl$ given
earlier
together with {\bf V1}--{\bf V4} and
{\bf Nec$_V$},
except that {\bf A5$_n$} is replaced by {\bf A5$_n'$}.
AX$'$ is the
axiom system that provides what we claim is the desired
generalization of Levesque's axioms to the multi-agent case.
In particular, {\bf A5$_n'$} is the appropriate generalization
of {\bf A5}.  The question is, of course, whether there is a
semantics for which this is a complete axiomatization.  We now
provide one, in the spirit of the canonical-model construction
of Section~\ref{canonical-model},
except that, in the spirit of the extended situations of
Section~\ref{review}, we do not attempt to make the set of worlds
used for evaluating $L_i$ and $N_i$ disjoint.

\begin{sloppypar}
Let $\onlp$ be the extension of $\onln$ to include the modal
operator $\Vall$.  For the remainder of this section, when we say
``consistent'', we mean consistent with the axiom system
AX$'$.
We define the \newl{extended canonical model}, denoted
$M^e = (W^e, \pi^e, \K_1^e, \ldots, \K_n^e,\N_1^e,\ldots,\N_n^e)$,
as follows:
\end{sloppypar}

\begin{itemize}
\item $W^e$ consists of the
maximal consistent sets of formulas in $\onlp$.
\item For all primitive propositions
$p$ and  $w \in W^e$, we have $\pi^e(w)(p) = \true$
  iff $p\in w$.
\item $(w,w') \in   \K_i^e$ iff $w/L_i \subseteq w'$.
\item $(w,w') \in \N_i^e$ iff $w/N_i \subseteq w'$.
\end{itemize}

In this canonical model, the semantics for $L_i$ and $N_i$ is defined
in terms of the $\K_i^e$ and $\N_i^e$ relations, respectively:
$$
\begin{array}{l}
\mbox{$(M^e,w) \sat L_i \alpha$ \ \ if \ \ $(M^e,w') \sat \alpha$ for
all $w'$ such that $(w,w') \in \K_i^e$.}\\
\mbox{$(M^e,w) \sat N_i \alpha$ \ \
if \ \  $(M^e,w') \sat \alpha$ for all $w'$
such that $(w,w') \in \N_i^e$.}\end{array}
$$
We define the $\Vall$ operator so that it corresponds to validity
in the extended canonical model:
$$\mbox{$(M^e,w) \sat \Val{\alpha}$ if $(M^e,w') \sat \alpha$ for all
worlds $w'$ in $M^e$.}$$

We now want to show that every formula in a maximal consistent
set is satisfied at a
world in the
extended canonical model. To do this, we need
one preliminary result, showing
that $\Vall$ and $\Conn$ really correspond to
provability and consistency in this framework.
\pro\label{valok}
For every formula $\phi \in \onln$, if $\phi$ is provable then
so is $\Val{\phi}$, while if $\phi$ is not provable, then
$\neg \Val{\phi}$ is provable.
\epro

\prf By {\bf Nec$_V$}, it is clear that if $\phi$ is provable, so is
$\Val{\phi}$.  Thus, it remains
to show that if $\phi$ is not provable, then $\neg \Val{\phi}$ is.
Using {\bf V1}, it is easy to see that $\neg \Val{\phi}$ is provably
equivalent to $\Con{\neg \phi}$, so it suffices to show that if
$\phi$ is not provable---\ie if $\neg \phi$ is consistent---then
$\Con{\neg \phi}$ is provable.  We prove by induction on $\phi$
that if $\phi$ is consistent, then $\Con{\phi}$ is provable.

If $\phi$
is propositional, the result is immediate from {\bf V2}.  For the
general case,
we first use Lemma~\ref{normalform} to restrict attention to formulas
in the canonical form specified by the lemma.  Using standard modal
reasoning ({\bf V1} and {\bf Nec$_V$}) it is easy to show that
$\vdash \Con{\phi \lor \psi} \dimp (\Con{\phi} \lor \Con{\psi})$.
Thus, it suffices to restrict attention to a conjunction in the form
specified by the lemma.  It is easy to see that if the conjunction
is consistent, then each conjunct must be consistent.  Using
{\bf V4}, it is easy to see that we can restrict attention to
$i$-subjective formulas.  By applying Lemma~\ref{normalform}, we can
assume without loss of generality that we are dealing with a
consistent formula $\phi$ of the form
$L_i \alpha \land \neg L_i \neg \beta_1
\land \ldots \land \neg L_i \neg \beta_k \land N_i \gamma \land
\neg N_i \neg \delta_1 \land \ldots \land \neg N_i \neg \delta_m$,
where $\alpha, \beta_1, \ldots, \beta_k, \gamma, \delta_1, \ldots,
\delta_m$ are all $i$-objective.  We can also assume that
each of $\alpha \land \beta_i$, $i = 1, \ldots, k$ and
$\gamma \land \delta_j$, $j = 1, \ldots, m$ are consistent, for
otherwise we could easily show that $\phi$ is not consistent.
Finally, we can show that $\alpha \lor \gamma$ must be provable,
for if not, by applying {\bf A5$'_n$}, we can again show that
$\phi$ is not consistent.  We now apply the induction hypothesis
to prove the result.  \eprf

\cor\label{noval} Each formula in $\onlp$ is provably equivalent to a
formula in $\onln$.
\ecor

\prf We proceed by induction on the structure of formulas.  The only
nontrivial case is for formulas of the form $\Val{\phi}$.  By the
induction hypothesis, $\phi$ is provably equivalent to a formula
$\phi' \in \onln$.  By straightforward modal reasoning using {\bf V1}
and {\bf Nec$_V$}, we can show that $\Val{\phi}$ is provably
equivalent to $\Val{\phi'}$.  By Proposition~\ref{valok}, $\Val{\phi'}$
is provably equivalent to either $\true$ or $\false$, depending
on whether $\phi'$ is provable. \eprf


Using standard modal logic techniques, we can now prove the following
result.
\thm\label{complete3}
$M^e$ is a $\Wax$ structure
(that is, $\K_i^e$ and $\N_i^e$ are Euclidean and transitive).
Moreover, for each world $w\in W^e$, we have
$(M^e,w) \sat \alpha$ iff $\alpha
\in w$.
\ethm

\prf
We leave it to the reader to check that the definition of $\K_i^e$
guarantees that $M^e$ is a $\Wax$ structure.  Given
Corollary~\ref{noval}, which allows us to restrict attention
to $\alpha \in \onln$, the proof that
$(M^e,w) \sat \alpha$ iff $\alpha \in w$
is completely straight forward and
follows the same lines as the usual proofs dealing with
canonical models (see, for example, \cite{Chellas,HM2}).
\eprf

\deleted{
\prf We leave it to the reader to check that the definition of $\K_i^e$
guarantees that $M^e$ is a $\Wax$ structure.  We now
prove that $(M^e,(\Gamma,x)) \sat \alpha$ iff $\alpha \in \Gamma$.
By Corollary~\ref{noval}, it suffices to restrict attention to
$\alpha \in \onln$.  We proceed by induction on structure.  All the
cases follow the same lines as the standard proofs dealing with
canonical models (see, for example, \cite{Chellas,HM2}), except that
for formulas of the form $N_i \phi$.  We deal with this case
as follows.

First suppose $N_i \phi \in \Gamma$.  It clearly suffices to deal with
formulas in the canonical form described in Lemma~\ref{normalform}, so
we can assume without loss of generality that $\phi$ is $i$-objective.
Let $w = (\Gamma,x)$, and
suppose that $w' = (\Gamma',x') \approx_i w$ and $w' \notin \K_i^e(w)$.
There are two cases to consider.  First suppose
that for some $i$-objective $\phi' \in \onln$,
we have $L_i \phi' \in \Gamma$ and $\neg
\phi' \in \Gamma'$.  Since $N_i \phi \land L_i \phi' \in \Gamma$,
it follows from {\bf A5$_n$} that $\Val{\phi \lor \phi'}$.  From
Proposition~\ref{valok}, it follows that $\phi \lor \phi'$, or
equivalently, $\neg \phi' \rimp \phi$, is provable.  Since $\neg \phi'
\in \Gamma'$, it follows that $\phi \in \Gamma'$ and, by the induction
assumption, that $(M^e,(\Gamma',x')) \sat \phi$.
If the first case does not apply, the definition of $\K_i^e$
guarantees that it must be the case that $x'
= N$ and $\Gamma/N_i \subseteq \Gamma'$.  But this means that $\phi' \in
\Gamma'$, and again by the inductive hypothesis we have that
$(M^e,(\Gamma',x')) \sat \phi$.

Now suppose that $\neg N_i \phi' \in \Gamma$.  We claim that $\Gamma/N_i
\union \{\neg \phi'\}$ must be consistent, for if not,
there must be a finite subset $\Delta$ of
$\Gamma/N_i$ such that $\Delta \rimp \phi'$ is provable, and standard
modal logic arguments show that $N_i \phi'$ must be in $\Gamma$,
contradicting the consistency of $\Gamma$.  It follows that there is
some maximal consistent set $\Gamma'$ containing $\Gamma/N_i \union
\{\neg \phi\}$.  Let $w' = (\Gamma',N)$.  By the inductive hypothesis,
we have $(M^e,w') \sat \neg \phi'$.   Notice that if $\psi$ is an
$i$-subjective formula
in $\Gamma$, then by {\bf A4$_n$}, $N_i \psi \in \Gamma$, so $\psi \in
\Gamma/N_i$.  Thus, $\Gamma$ and $\Gamma'$ agree on all $i$-subjective
formulas, so $w \approx_i w'$.  Moreover, since $\Gamma/N_i \subseteq
\Gamma'$, the definition of $\K_i^e$ guarantees that $w' \notin
\K_i^e(w)$.  Thus, $(M^e,w) \sat \neg N_i \phi'$, as desired.  \eprf
} 

We say that $\alpha$ is \newl{e-valid}, denoted $\sat^e \alpha$, if
$M^e \sat \alpha$, that is, if
$(M^e,w) \sat
\alpha$ for all worlds $w \in W^e$.  The following result is immediate
from Theorem~\ref{complete3}.
\cor $\sat^e \alpha$ iff $AX' \vdash \alpha$.  \ecor
Thus, AX$'$ is a sound and complete axiomatization of $\onlp$ with
respect to the $\sat^e$ semantics.

While AX$'$ is sufficient for our purposes, it comes at the expense
of having to explicitly axiomatize validity as part of the logic
itself.
While we view the ability to axiomatize validity and
satisfiability within the logic as a feature in our approach, it is
reasonable to ask whether it is really necessary.
 One of the anonymous referees suggested to us the following
interesting variant,
which may avoid this complication, although
at the expense of an infinite
number of axiom schemas.

First we define an infinite sequence of languages
\onlnk\ for $k = 0,1,2\ldots$:

\begin{itemize}
\item \onlnzero = \onlnminus
\item \onlnkplusone = $\{\alpha\;|\;$\parbox[t]{4.5in}{$\alpha$ is
a Boolean combination of formulas of \onlnk\ together with\\
formulas of the form $L_i\alpha$ or $N_i\alpha$ for $\alpha\in\onlnk$$\}$}
\end{itemize}

Roughly, each language adds another level of nestings of only knowing
with varying agent indices. For example, \onlnkplusone\ contains the formula
$O_{i_0}O_{i_1}\ldots O_{i_{k+1}} p$, where $i_j \ne i_{j+1}$, something
that cannot be expressed in \onlnk.

Let \AXnew\ consist of the the axioms {\bf A1$_n$}--{\bf A4$_n$}
as before together with the following set of axioms
\begin{quote}
\Afivenewkplusone. $N_i\alpha \rimp \lnot \L_i\alpha$, where
$\alpha$ is an i-objective \onlnk\ formula which is consistent
\wrt {\bf A1$_n$}--{\bf A4$_n$},\Afivenewone, \Afivenewtwo,$\ldots$ \Afivenewk.
\end{quote}

It is not hard to show that the axioms are sound with respect
to the semantics. Whether they are also complete remains an open
problem.

Apart from the question of axiomatization, how does the $\sat^e$ semantics
compare to our earlier two?  Clearly, they differ.  It is easy to see that the
formula $O_i \neg O_j p$, which was not satisfiable under $\mdcm$, is
satisfiable under $\sat^e$.  In addition, the formula $N_i \neg O_j p \land L_i
\neg O_j p$, which is satisfiable under $\sat'$, is not satisfiable under
$\sat^e$.  In both cases, it seems that the behavior of $\sat^e$ is more
appropriate.  On the other hand, all three semantics agree in the case where our
intuitions are strongest, $\onlnminus$.  Since the axiom system AX characterizes
how our earlier two semantics deal with $\onlnminus$,
this is shown by the following result.

\thm If $\phi \in \onlnminus$, then $AX \vdash \phi$ iff
$AX' \vdash \phi$.  \ethm

\prf It is easy to see that each axiom of AX is sound in AX$'$.  It
follows that $\mbox{AX} \vdash \phi$ implies
AX$' \vdash \phi$.  For the converse, it suffices to show that if
$\phi \in \onlnminus$ is consistent with AX, then it is also
consistent with AX$'$, \ie that $\Con{\phi}$ holds.
We show this by induction
on the structure of $\phi$, much in the same way we proved
Proposition~\ref{valok}.  We can assume without loss of generality
that $\phi$ is a conjunction in the normal form described
Lemma~\ref{normalform}.  It is easy to see that if we can deal with the
case that $\phi$ is an $i$-subjective formula, then we can deal with
arbitrary $\phi$ by repeated applications of {\bf V4} followed
by an application of {\bf V2}.  Thus, suppose that $\phi $ is an
$i$-subjective formula which is consistent with AX.  We can assume that
$\phi$ is of the form
$L_i \alpha \land \neg L_i \neg \beta_1
\land \ldots \land \neg L_i \neg \beta_k \land N_i \gamma \land
\neg N_i \neg \delta_1 \land \ldots \land \neg N_i \neg \delta_m$.
We must have  that $\alpha \land \beta_j$ is consistent for
$j = 1, \ldots, k$, and that $\gamma \land \delta_l$ is AX-consistent
for
$l = 1, \ldots , m$, for otherwise $\phi$ would not be AX-consistent.
Similarly, by Lemma~\ref{LN}, we must have that $\alpha \lor \gamma$ is
$\Wax$-provable, otherwise $\phi$ would not be AX-consistent.  We can
now apply {\bf V3} and the inductive hypothesis to show that $\alpha$ is
AX$'$-consistent. \eprf

Thus, we maintain all the benefits of the earlier semantics with this
approach.  Moreover,
while it is just as intractable to compute whether $(M^e,w) \sat \alpha$
for a particular world $w$ in the extended canonical model as it was in
all our other approaches, we can show that
the validity problem for this
logic is no harder than that for $\Wax$ alone.  It is PSPACE-complete.

\thm The problem of deciding if $AX' \vdash \phi$ is PSPACE-complete.
\ethm

\prf PSPACE hardness follows from the PSPACE hardness of $\Wax$
\cite{HM2}.\footnote{The result in \cite{HM2} is proved only for $\Dax$,
but the same proof applies to $\Wax$.}  We sketch the proof of the upper
bound.  First of all, observe that it suffices to deal with the case
that $\phi$ is in $\onln$, since we can then apply the arguments of
Corollary~\ref{noval} to remove all occurrences of $\Vall$ from inside
out.  We consider the dual problem of consistency.  Thus, we want to
check if $\Con{\alpha}$ holds.
The first step is to convert
$\alpha$ to the normal form of Lemma~\ref{normalform}.  Observe that
$\alpha$ is consistent iff at least one of the disjuncts is consistent.
Although the conversion to normal form may result in exponentially many
disjuncts, each one is no longer than $\alpha$.  Thus, we deal with them
one by one, without ever writing down the full disjunction.  It suffices
to show that we can decide if each disjunct is consistent in polynomial
space, since we can then erase all the work and start over for the next
disjunct (with a little space necessary for bookkeeping).  We now
proceed much as in the proof of Proposition~\ref{valok}.  By applying
{\bf V4} repeatedly and then {\bf V2} (as in the previous theorem), it suffices to
deal with $i$-subjective formulas.  We then apply {\bf V3} to get simpler
formulas, and repeat the procedure.
We remark that this gives another PSPACE decision procedure for $\Wax$,
quite different from that presented in \cite{HM2}. \eprf

To what extent do the three properties we have been focusing on hold under the
$\sat^e$ semantics?  Suppose we take the conceivable states from $i$'s point of
view to be the maximal consistent sets of $i$-objective formulas with respect to
AX$'$, or equivalently, the set of $i$-objective formulas true at some world in
$M^e$.  Let \newl{$\objei(M^e,w)$} consist of all the $i$-objective formulas
true at world $w$ in the extended canonical model (under the $\sat^e$
semantics) and let \newl{$\Objei(M^e,w)$} = $\{\objei(M^e,w') \;|\; w' \in
\K_i^e(w)\}$.  It is easy to see that the first two properties we isolated hold
under this interpretation of conceivable state.  However, it is quite possible
that the ``possible states'' at a world $(M^c,w)$, that is, $\Objei(M^c,w)$, and
the ``impossible states'', that is, $\{\objei(M^c,w') \;|\; w' \approx_i w, w
\notin \K_i^e(w)\}$ are not disjoint.

Interestingly, this semantics does not satisfy the third
property we isolated.  Not all subsets of conceivable states
arise as the set of possible states at some situation $(M^e,w)$.
A proof analogous to that of Lemma~\ref{lc} shows that $\Objei(M,w)$
is always limit closed.
In the canonical model approach, limit closure prevents an agent from
considering certain desirable sets of states. Now this is no longer
the case, despite limit closure. Roughly speaking,
we avoid problems by having in a precise sense ``enough'' possibilities.
More precisely, given any consistent $i$-objective
formula $\alpha$, it is possible for agent $i$ to only
know $\alpha$
by virtue of considering all maximal sets of $i$-objective
formulas which contain $\alpha$.
Thus, as we
suggested earlier,
the third property
we isolated in the beginning turns out to be somewhat too strong---%
it is sufficient but not necessary once we allow for enough possibilities.

\section{Multi-Agent Nonmonotonic Reasoning}\label{apps}
In this section, we demonstrate that the logic developed in
Section~\ref{synthesis} captures multi-agent
autoepistemic reasoning in a reasonable way.
We do this in two ways.  First we show by example that
the logic can be used to derive some reasonable
nonmonotonic inferences in a multi-agent context.
We then show that the logic can be used to extend
the definitions of stable sets and stable expansions
originally developed for single agent autoepistemic logic
to the multi-agent setting.

\subsection{Formal Derivations of Nonmonotonic Inferences}
\label{nonmon}
In this section, we provide two examples of how the logic can be used
for nonmonotonic reasoning.

\xam
Let $p$ be agent $i$'s secret and suppose $i$ makes the following
assumption: unless I know that $j$ knows my secret assume that $j$ does
not know it. We can prove that if this assumption is all $i$
believes then he indeed believes that $j$ does not know his secret.
Formally, we can show
$$\vdash  O_i(\lnot L_i L_j p \rimp \lnot  L_j p)
             \rimp  L_i\lnot L_j p.%
\footnote{Note that if we replace $ L_j p$ by
$p$ we obtain regular single-agent autoepistemic reasoning.}$$
A formal derivation of this theorem can be obtained as follows. Let
\mbox{$\alpha = \lnot L_i L_j p \rimp \lnot  L_j p$.} The justifications
in the following derivation indicate which axioms or previous
derivations have been used to derive the current line. PL
or K45$_n$ indicate that reasoning in either standard propositional logic
or K45$_n$, which are subsumed by AX$'$, is used without further analysis.

\begin{tabular}{lll}
1. & $ O_i\alpha \rimp  L_i\alpha$ & PL\\
2. & $ O_i\alpha \rimp N_i\lnot\alpha$ & PL\\
3. & $( L_i\alpha \land \lnot L_i L_j p) \rimp  L_i\lnot L_j p$ &
\mbox{K45$_n$} \\
4. & $N_i\lnot\alpha \rimp (N_i\lnot L_i L_j p \land N_i L_j p)$ &
\mbox{K45$_n$}\\
5. & $\Con{p}$ & {\bf V2}\\
6. & $\Con{p} \rimp \Con{\lnot L_j p}$ & {\bf V3}\\
7. & $\Con{\lnot L_j p}$ & {\bf V1}, PL\\
8. & $\Con{\lnot L_j p} \rimp (N_i L_j p \rimp \lnot  L_i L_j p)$ & {\bf A5$'_n$}\\
9. & $N_i L_j p \rimp \lnot  L_i L_j p$ & {\em PL}\\
10. & $ O_i\alpha \rimp \lnot L_i L_j p$ & 2; 4; 9; PL\\
11. & $ O_i\alpha \rimp  L_i\lnot L_j p$ & 1; 3; 10; PL\\
\end{tabular}
\ \\[2ex]
To see that $i$'s beliefs may evolve nonmonotonically given that $i$ knows
only
$\alpha$, assume that $i$ finds out that $j$ has found out about the secret.
Then $i$'s belief that $j$ does not believe the secret will be retracted.
In fact, $i$ will believe that $j$ does believe the secret.
Formally, we can show
$$\vdash  O_i( L_j p\land \alpha)              \rimp L_i L_j p.$$
Notice that the logic itself is a regular monotonic logic; the
nonmonotonicity of agent $i$'s beliefs is hidden within the
$ O_i$-operator.

All the formulas that appear in the proof above are in $\onlnminus$.
Thus, we could have used the somewhat simpler proof theory of
Section~\ref{prooftheory}. To obtain an example where we need
the full power of AX$'$, simply replace $L_j p$ by $O_j p$, that
is $i$ now uses the default that unless he knows that $j$ only knows
$p$, then he assumes that $j$ does not only know $p$. In other words,
$i$ (prudently) makes rather cautious assumptions about $i$'s epistemic state
and assumes that $i$ usually knows more than just $p$. The proof
is very similar to the one above.
The only difference is that we now have to establish
that $\Con{\lnot O_j p}$ is provable, which is straightforward.
\exam
\xam
\label{proof-of-$i$-knows-that-j-knows-that-Tweety-flies}
Now let $p$ stand for
``Tweety flies''.%
\footnote{Regular readers of papers on nonmonotonic logic will no doubt
be gratified to see Tweety's reappearance.}
We want to show that if $j$ knows that all $i$ knows about Tweety is that
by default it flies,
then $j$ knows that $i$ believes that Tweety flies.
As before, we capture the fact that all $i$ believes is that, by
default, Tweety flies, by saying
that all $i$ believes is that, unless $i$ believes that Tweety does not
fly, then Tweety flies.   Thus, we want to show
$$\vdash L_j O_i (\neg L_i \neg p \rimp p) \rimp L_j L_i p.$$
We proceed as follows:

\begin{tabular}{lll}
1. & $O_i(\lnot L_i\lnot p \rimp p) \rimp  L_i p$ & as above, with
$L_j p$ replaced by $\neg p$\\
2. & $L_j (O_i(\lnot L_i\lnot p \rimp p) \rimp L_i p)$ & 1; {\bf Nec$_n$}
\\
3. & $L_j O_i(\lnot L_i\lnot p \rimp p) \rimp  L_j L_i p$ & 2;
\mbox{K45$_n$}
\end{tabular}
\ \\[2ex]
In a sense, $i$ is able to reason about $j$'s ability
to reason nonmonotonically essentially by simulating $j$'s reasoning pattern.
\exam
A situation where $i$ knows that {\em all\/} $j$
knows is $\alpha$ seems hardly attainable in practice, since
an agent usually has at best incomplete information about another agent's
beliefs. It would seem much more
reasonable if we could say that $i$ knows that $\alpha$ is all $j$ knows about
{\em some relevant subject}, say Tweety. This issue is dealt
with in~\cite{Lakemeyer93b}, where the canonical-model approach is extended
to allow statements of the kind that all agent $i$ knows about $x$ is $y$.
It is shown that the forms of nonmonotonic reasoning just described,
when restricted to $\onlnminus$,
go through just as well with the weaker notion of only knowing about.
%

\subsection{$i$-Stable Sets and $i$-Stable Expansions}
\label{stable-sets}
Single-agent autoepistemic logic was
developed by Moore \citeyear{Moore85} using the concepts of
{\em stable sets} and {\em stable expansions}.
Levesque proved that there is a close relationship between stable sets
and only-knowing in the single-agent case.  Here we prove an analogous
relationship for the multi-agent case.  We first need to define a
multi-agent analogue of stable sets.

In the single-agent case, it is well known that a stable set is a
complete set of formulas that agent $i$ could know in some situation;
that is, a set $S$ is stable if and only if there is a situation $(W,w)$
such that $S = \{\alpha\;|\;(W,w) \sat L \alpha\}$.  This is the intuition
that we want to extend to the multi-agent setting, where the underlying language
is now $\onln$. First we define logical consequence in the extended canonical
model
in the usual way:
If $\Gamma$ is a set of formulas, we write $M^e \sat \Gamma$ if $M^e
\sat \gamma'$ for each formula $\gamma' \in \Gamma$.
We say that $\gamma$ is an {\em \econsequence\/} of $\Gamma$, and
write $\Gamma\conse\gamma$, exactly if
$M^e\sat\Gamma$ implies $M^e \sat \gamma$.
\dfn
Let $\Gamma$ be a set of formulas in $\onln$.
$\Gamma$ is called {\em $i$-stable\/} iff
\begin{enumerate}
  \item[(a)] if $\Gamma\conse\gamma$ then $\gamma\in\Gamma$,
  \item[(b)] if $\alpha\in\Gamma$ then $ L_i\alpha\in\Gamma$,
  \item[(c)]
  if $\alpha\not\in\Gamma$ then $\lnot L_i\alpha\in\Gamma$. \bbox
\end{enumerate}
\end{definition}

Note that the only difference between $i$-stable sets and the original
definition
of stable sets is in condition (a), which requires $i$-stable sets to be
closed under \econsequence\
instead of tautological consequence
(\ie logical consequence in propositional logic)
as in the single-agent case.
Using \econsequence\ rather than tautological consequence makes no
difference in the single-agent case; in the multi-agent case it does.
Intuitively, we want to allow the agents to use \econsequence\  here to
capture the intuition that it is common knowledge that all agents
are perfect reasoners under the extended canonical model semantics.
For example, if agent $i$ believes $\lnot L_j p$
for a different agent $j$, then we want him to also believe
$N_j\lnot L_j p$.  To do this, we need to close off under \econsequence.
\deleted{
It is also interesting to note that Halpern and Moses~\cite{HalpernMoses84}
proposed a very similar definition of stability in the n-agent case.
The difference is that they require stable sets to be consistent and
to contain all valid formulas of the logic {\em KT5$_n$}, also called {\em S5$_n$}.
} 

The next theorem shows that $i$-stable sets do indeed satisfy the
intuitive requirement.  We define an \newl{$i$-epistemic\/} state to be
a set $\Gamma$ of $\onln$-formulas such that for some situation
$(M^e,w)$ in the extended canonical model,
$\Gamma = \{\alpha\in\onln\;|\; (M^e,w) \sat L_i \alpha\}$;
in this case, we say that $\Gamma$ is the $i$-epistemic situation
corresponding to $(M^e,w)$.
For a set of $\onln$-formulas $\Gamma$, let $\Gammabar = \{\gamma\;|$
$\gamma\in\onln$ and $\gamma\not\in\Gamma\}$,
$ L_i\Gamma = \{ L_i\gamma\;|\; \gamma\in\Gamma\}$, and
$\lnot L_i\Gammabar = \{\lnot L_i\gamma\;|\;
\gamma\in\Gammabar\}$.

%
\thm
Let $\Gamma$ be a set of $\onln$-formulas. $\Gamma$ is $i$-stable iff
$\Gamma$ is an $i$-epistemic state.
\ethm
\prf It is straightforward to show that every $i$-epistemic state is
$i$-stable. To show the converse, let $\Gamma$ be $i$-stable. We need to
show that
it is also an $i$-epistemic state. Certainly $ L_i\Gamma$ is consistent.
If $\Gamma$ contains all $\onln$-formulas, that is, if agent $i$ is
inconsistent,
then let $(M^e,w)$ be a situation where $\K_i^e(w) = \emptyset$; such a
situation clearly exists. Then $\Gamma$ is the
$i$-epistemic situation corresponding to $(M^e,w)$.
If $\Gamma$ is a proper subset of
the $\onln$-formulas,
then $\Gamma$ must be consistent. (If $\Gamma$ were inconsistent, by the
first property of stable sets, $\Gamma$ would contain all formulas.)
In particular, this means that $p \land \neg p \notin \Gamma$ for a
primitive proposition $p$.
The second property of stable sets guarantees that $L_i \Gamma \subseteq
\Gamma$, while the third guarantees that $\neg L_i (p \land \neg p) \in
\Gamma$.  Since $\Gamma$ is consistent, so is $L_i \Gamma \union \{ \neg
L_i (p \land \neg p)\}$.
$W^e$ consists of all the maximal consistent sets (with respect to
AX$'$); thus, there must be some $w^* \in W^e$ that contains
$ L_i\Gamma\union \{\lnot L_i(p\land\lnot p)\}$.
We claim that $\Gamma$ is the $i$-epistemic state corresponding to
$(M^e,w^*)$.  Thus, we must show that $\phi \in \Gamma$ iff $(M^e, w^*)
\sat L_i \phi$.
To see this, first suppose that $\phi \in \Gamma$.  Thus, $L_i \phi \in
L_i \Gamma$.  By construction, $w^*$ contains $L_i \Gamma$.  By
Theorem~\ref{complete3}, we have that $(M^e,w^*) \sat L_i \phi$.
On the other hand, if $\phi \notin \Gamma$, then, since $\Gamma$ is
$i$-stable, we have that $\neg L_i \phi \in \Gamma$.  By the previous
argument, it follows that $(M^e,w^*) \sat L_i \neg L_i \phi$.
{F}rom ${\bf A4}_n$, it follows that $(M^e,w^*) \sat \neg L_i \phi$, and
hence $(M^e,w^*) \not\sat L_i \phi$.   This proves the claim.
\eprf

Moore \citeyear{Moore85}
defined the notion of a {\em stable expansion\/} of a
set $A$ of formulas in the single-agent case.  Intuitively, a
stable expansion of $A$ is a stable set containing $A$ all of whose
formulas can be justified, given $A$ and the formulas believed in that
stable set.  Further discussion and justification of the
notion of stable expansion can be found in \cite{Hal13,Moore85}.
Rather than
discussing this here, we go directly to our multi-agent generalization
of Moore's notion.

\dfn
Let $A$ be a set of $\onln$-formulas.
$\Gamma$ is called an \newl{$i$-stable expansion\/} of $A$ iff
$\Gamma = \{ \gamma\in\onln \;|\;
A\union L_i\Gamma \union\lnot L_i\Gammabar\conse\gamma\}$.
\edfn
%
It is easy to see that an $i$-stable expansion is an $i$-stable set.  The
definition of $i$-stable expansions looks exactly like Moore's definition
of stable expansions except that we again use \econsequence\ instead of
tautological consequence.  As in the case of stable sets, this is necessary
to capture the fact that it is common knowledge that all agents can do
reasoning under the extended canonical model semantics.

We now generalize a result of Levesque's \citeyear{Lev5} (who
proved it for the single-agent case), showing that the $i$-stable
expansions
of a formula $\alpha$ correspond precisely to the different
situations where $i$ only knows $\alpha$.
We first need a lemma.

\lem
\label{e-states-decide-all-i-subj-form}
Let $(M^e,w)$ be a situation with
\[\Sigma = \{L_i\gamma \;|\;
(M^e,w)\md L_i\gamma\}
\union \{\lnot L_i\gamma \;|\; (M^e,w)\md \lnot L_i\gamma\}.
\]
For any $\alpha$,
there is an $i$-objective formula $\alpha^*$ such that
\begin{enumerate}
\item[(a)] $\Sigma \conse \alpha\dimp \alpha^*$,
\item[(b)] $(M^e,w) \sat (L_i \alpha \dimp L_i\alpha^*) \land (N_i \alpha
\dimp N_i\alpha^*)$.
\end{enumerate}
\elem
\prf
Given a formula $\phi$, we say that a subformula
$L_i \psi$ of $\phi$ occurs at {\em top level\/} if it is not in the
scope of any modal operators.
Let $\alpha^*$ be the result of replacing each top-level subformula of
$\alpha$ of the form $L_i \gamma$ (\respc $N_i \gamma$) by $\true$
if $\Sigma\conse L_i\gamma$
(\respc $\Sigma\conse N_i\gamma$) and by $\false$
otherwise.  Clearly $\alpha^*$ is $i$-objective.  Moreover, a trivial
argument by induction on the structure of $\alpha$ shows that
$\Sigma \sat \alpha \dimp \alpha^*$: If $\alpha$ is a primitive
proposition then $\alpha^* = \alpha$; if $\alpha$ is of the form
$\alpha_1 \land \alpha_2$ or $\neg \alpha'$, then the result follows
easily by the induction hypothesis; if $\alpha$ is of the form $L_j
\beta$ for $j \ne i$, then $\alpha = \alpha^*$, since $\alpha$ has no
top-level subformulas of the form $L_i \phi$; finally, if $\alpha$ is
of the form $L_i \beta$, then $\alpha^*$ is either $\true$ or $\false$,
depending on whether $L_i \beta$ is in $\Sigma$.  Since either $L_i
\beta$ or $\neg L_i \beta$ must be in $\Sigma$, the result is immediate
in this case too.  Part (b) follows immediately from part (a), since if
$w' \approx_i w$, we must have $(M^e,w') \sat \Sigma$, so $(M^e,w')
\sat \alpha \dimp \alpha^*$. \eprf


\thm 
\label{O-vs-stable-expansions}
Let $w$ be a world in the extended canonical model and let
$\Gamma$ be the $i$-epistemic state corresponding to $(M^e,w)$.
Then, for every $\onln$-formula $\alpha$, we have that
$(M^e,w)\md O_i\alpha$ iff
      (a) $\Gamma$ is an $i$-stable expansion of $\{\alpha\}$ and
(b) $\K_i^e(w)$
         and $\N_i^e(w)$ are disjoint.
\ethm
\prf
Let $\Sigma = L_i \Gamma \union \neg L_i \Gammabar$. To prove the ``only if''
direction, suppose $(M^e,w)\md O_i\alpha$. The disjointness of $\K_i^e(w)$ and
$\N_i^e(w)$ follows immediately from the fact that $(M^e,w)\md L_i\alpha\land
N_i\lnot\alpha$.  To prove that $\Gamma$ is an $i$-stable expansion of
$\{\alpha\}$, it suffices to show that for all $\onln$-formulas $\beta$, we have
$(M^e,w)\md L_i\beta\ {\rm iff}\ \{\alpha\}\union\Sigma \conse \beta$.

First suppose that $\{\alpha\}\union\Sigma \conse \beta$.
Since $(M^e,w)\md \{O_i\alpha\} \union \Sigma$, and every formula in $\Sigma$ is
of the form $L_i \gamma$ or $\neg L_i \gamma$, it easily follows that $(M^e,w)
\md L_i \alpha$ and  $(M^e,w)\md L_i \Sigma$.  Hence, $(M^e,w') \md \alpha$
and $(M^e,w') \md \Sigma$
for every $w'\in \K_i^e(w)$.
It follows that $(M^e,w')\md\beta$ and, therefore, $(M^e,w)\md L_i\beta$.

For the converse,
suppose that $(M^e,w)\md L_i\beta$.  We want to show that
$\{\alpha\} \union \Sigma \conse \beta$.
By Lemma~\ref{e-states-decide-all-i-subj-form}, we can assume without
loss of generality that $\beta$ is $i$-objective.  (For if not, we can
replace $\beta$ by an $i$-objective $\beta^*$ such that $\Sigma \conse
\beta \dimp \beta^*$ and $(M^e,w) \sat L_i \beta \dimp L_i \beta^*$,
prove the result for $\beta^*$, and conclude that it holds for $\beta$
as well.)  To show that $\{\alpha\} \union \Sigma \conse \beta$, we must
show that for all worlds $w'$ such that
$(M^e,w')\md \{\alpha\}\union\Sigma$, we have $(M^e,w') \sat \beta$.
By Lemma~\ref{e-states-decide-all-i-subj-form}, there is an
$i$-objective formula $\alpha^*$ such that $(M^e,w) \sat O\alpha \dimp
O\alpha^*$ and $\Sigma \conse \alpha \dimp \alpha^*$.  Thus, $(M^e,w)
\sat N_i \alpha^* \land L_i \beta$.  By the arguments of Lemma~\ref{LN}
(which apply without change to the extended canonical model semantics),
it must be the case that $\conse \alpha^* \rimp \beta$.  Since $\Sigma
\conse \alpha \dimp \alpha^*$, it follows that $\{\alpha\} \union \Sigma
\conse \beta$.
%
Since $(M^e,w')\md
\{\alpha\}\union\Sigma$ by assumption, we have that $(M^e,w')\md\beta$,
as desired.

To prove the ``if'' direction of the theorem, suppose that
$\K_i^e(w)$ and $\N_i^e(w)$ are disjoint and that for all
$\onln$-formulas $\beta$, we have $(M^e,w)\md L_i\beta\ {\rm iff}\
\{\alpha\}\union\Sigma \conse \beta$. We need to show that $(M^e,w)\md
O_i\alpha$, that is, $(M^e,w)\md L_i\alpha\land N_i\lnot\alpha$.

Since $\{\alpha\}\union\Sigma \conse \alpha$, the fact that
$(M^e,w)\md L_i\alpha$ follows immediately.
To prove that $(M^e,w)\md N_i\lnot\alpha$, let $w'\in\N_i^e(w)$ and
assume, to the contrary,
that $(M^e,w')\md \alpha$. Since $w'\in\N_i^e(w)$, it follows that
$\Sigma\subseteq w'$. Hence $w/L_i \subseteq w'$, from
which $w'\in \K_i^e(w)$ follows, contradicting the assumption that
$\K_i^e(w)$ and $\N_i^e(w)$ are disjoint.
\eprf

%

\section{Conclusion}\label{discussion}
We have provided three semantics for multi-agent only-knowing.
All agree on the subset $\onlnminus$, but they differ on formulas
involving nested $N_i$'s.  Although
a case can be made
that
the $\sat^e$ semantics comes closest to capturing our intuitions
for ``knowing at most'', our intuitions beyond $\onlnminus$ are not
well grounded.  It would certainly help to have more compelling
semantics
corresponding to AX$'$.

On the other hand, it can be argued that semantics does not play
quite as crucial a role when dealing with knowing at most as in
other cases.  The reason is that the structures we must deal with,
in general, have uncountably many worlds.  For example, whichever
of the three semantics we use, there must be uncountably many
worlds $i$-accessible from a situation $(M,w)$ satisfying
$O_i p$, at least one for every $i$-set that includes $p$.
To the extent that we are interested in proof theory, the
proof theory associated with $\sat^e$, characterized by
the axiom system AX$'$, seems quite natural.  The fact
that the validity problem is no harder in this setting
than that for $\Wax$ adds further support to its usefulness.
Of course, as we suggested above, rather than only knowing,
it seems more appropriate to reason about only knowing about a certain
topic.  Lakemeyer \citeyear{Lakemeyer93b} provides a semantics
for only knowing about, using the canonical-model approach.
It would be interesting to see if this can also be done using
the other approaches we have explored here.

\bibliographystyle{plain}
\bibliography{z}
\end{document}